\pdfoutput=1

\documentclass[11pt]{article}
\usepackage{amssymb}
\usepackage{afterpage}
\usepackage[ruled,vlined]{algorithm2e}
\usepackage{algpseudocode}
\usepackage{amsmath}
\usepackage[final]{acl}
\usepackage{subcaption}
\usepackage{booktabs} 
\usepackage{siunitx}
\usepackage{multicol}
\usepackage{multirow} 
\usepackage{times}
\usepackage{latexsym}
\usepackage{array}
\usepackage{setspace}
\usepackage{graphicx} 
\usepackage{float}
\algnewcommand\algorithmicinput{\textbf{Input:}}
\algnewcommand\algorithmicoutput{\textbf{Output:}}
\algnewcommand\Input{\item[\algorithmicinput]}
\algnewcommand\Output{\item[\algorithmicoutput]}

\usepackage[T1]{fontenc}

\usepackage[utf8]{inputenc}

\usepackage{microtype}

\usepackage{inconsolata}

\usepackage{graphicx}
\usepackage{enumitem}
\usepackage{colortbl}
\definecolor{em}{gray}{0.9}
\newcommand{\cem}{\cellcolor{em}}

\definecolor{darkpurple}{RGB}{160, 0, 160}
\definecolor{darkred}{RGB}{160, 0, 0}
\definecolor{darkblue}{RGB}{0, 0, 160}
\definecolor{mygreen}{RGB}{62,123,39}
\definecolor{deepgreen}{RGB}{0,80,0}
\hypersetup{
    colorlinks,
    linkcolor=darkpurple,
    anchorcolor=blue,
    citecolor=mygreen
}

\title{SAE-SSV: Supervised Steering in Sparse Representation Spaces for Reliable Control of Language Models}

\author{
Zirui He$^1$\,\,
Mingyu Jin$^2$\,\,
Bo Shen$^1$\,\, 
Ali Payani$^3$\,\,
Yongfeng Zhang$^2$\,\, 
Mengnan Du$^1$\thanks{Corresponding author.}\quad\\
  $^1$NJIT
\;\;\;  $^2$Rutgers University\;\;\;  $^3$Cisco\;\;\; 
}

\begin{document}
\maketitle

\begin{abstract}

Large language models (LLMs) have demonstrated impressive capabilities in natural language understanding and generation, but controlling their behavior reliably remains challenging, especially in open-ended generation settings. This paper introduces \textbf{\textit{a novel supervised steering approach}} that operates in sparse, interpretable representation spaces. We employ \textit{sparse autoencoders (SAEs)} to obtain sparse latent representations that aim to \textit{disentangle semantic attributes from model activations}. Then we train linear classifiers to identify a small subspace of task-relevant dimensions in latent representations. Finally, we learn supervised steering vectors constrained to this subspace, optimized to align with target behaviors. Experiments across sentiment, truthfulness, and politics polarity steering tasks with multiple LLMs demonstrate that our supervised steering vectors achieve higher success rates with \textit{minimal degradation in generation quality} compared to existing methods. Further analysis reveals that 
\textit{a notably small subspace} is sufficient for effective steering, enabling more targeted and interpretable interventions. Our implementation is publicly available at \url{https://github.com/Ineedanamehere/SAE-SSV}.
\end{abstract}

\section{Introduction}
Large language models (LLMs) have demonstrated impressive capabilities across a wide range of natural language understanding and generation tasks~\cite{ouyang2022training, wei2022finetuned}. Yet, as language models' scale increases, achieving reliable and interpretable behavior control remains a fundamental challenge~\cite{zhao2024explainability, sharkey2025open}. One promising approach for controllable generation is steering, which manipulates internal model representations during inference to influence behaviors without modifying model parameters by retraining or finetuning~\cite{rimsky2024steering, turner2308steering, han2024word}.

Recent steering methods control LLM behavior by modifying internal activations at different points in the inference process: modifying residual stream activations~\cite{zou2023representation}, injecting latent directions learned from contrastive data~\cite{kleindessner2023efficient}, and applying interpretable feature vectors extracted from sparse autoencoders or linear classifiers~\cite{huben2024sparse, DBLP:journals/corr/abs-2502-16681}. These steering techniques offer a lightweight and modular means of behavior control, and have been applied to enforce stylistic consistency~\cite{wang2024steering}, mitigate social biases, and align LLM outputs with safety or fairness objectives~\cite{li2025fairsteer}. Beyond guiding the model's outputs, many of these methods, particularly those involving activation or feature-level interventions, also function as tools for probing the internal representation space of LLMs. This dual role has positioned them at the intersection of behavior control and mechanistic interpretability~\cite{zhao2025beyond, ferrando2025do}. Nevertheless, most evaluations of steering methods have focused on constrained tasks with easily measurable outputs such as multiple-choice question answering or sentiment binary classification, where control success can be directly quantified~\cite{zou2023representation, im2025unified}. Other recent works have applied steering in agentic~\cite{rahn2024controlling} and refusal-control~\cite{zhao2025adasteer} settings. Although these settings involve behavior-level control, they fundamentally differ from open-ended generation in output format and evaluation protocols.

Unlike classification or structured QA, the \emph{open-ended generation setting} requires LLMs to generate coherent and attribute-consistent text from scratch~\cite{li2023contrastive}. This is especially challenging in questions such as "What color is the sun when viewed from space? Briefly explain the reason." The model must not only produce a factually correct response but also structure it fluently without predefined options. This setting is central to real-world applications such as dialogue systems, creative writing, and factual content generation, yet steering methods often struggle in this regime~\cite{becker2024textgen}. Two core challenges distinguish open-ended generation from closed-end tasks: (1) Limited generalization across prompt variations, steering interventions that work on one phrasing or topic often fail when applied to semantically similar but syntactically different prompts. 
and (2) Generate quality degradation under strong control, intensifying the steering signal may improve direction alignment but often harms generation fluency, coherence, or factuality~\cite{zhou2024evaluating}. These difficulties point to a deeper issue in \textit{how steering vectors are typically constructed}. Many existing approaches rely on global heuristics, such as mean difference vectors or unsupervised projections~\cite{jorgensen2024improving, chalnev2024improving}. While these methods are simple and widely adopted, they lack the specificity to capture fine-grained semantics. Furthermore, they operate in dense, entangled activation spaces~\cite{huben2024sparse} and often fail to leverage supervision, leading to unstable or unintended behaviors under distributional shift.

To overcome these limitations, we propose \textit{\textbf{SAE Supervised Steering Vectors (SAE-SSV)}}, a framework that enables targeted and interpretable interventions by operating in a sparse, task-aligned subspace. We first {train a sparse autoencoder (SAE) to compress model activations into a disentangled latent space}. Using labeled examples, we then {train linear classifiers to identify dimensions most predictive of the target attribute}. Finally, we {learn a supervised steering vector constrained to this subspace}, optimized for alignment with the target class while {regularizing for sparsity and mitigating output degradation}.  By focusing only on task-relevant dimensions, our \textit{\textbf{SAE-SSV}} method addresses the trade-off between steering strength and generation quality that limits existing approaches.
Our contribution  can be summarized as follows:
\vspace{-5pt}
\begin{itemize}[leftmargin=*]\setlength\itemsep{-0.3em}
    \item  We propose \textit{SAE-SSV}, a supervised steering framework that constrains interventions to a sparse, task-relevant latent subspace identified via labeled data and sparse autoencoders. 
    \item Our method consistently outperforms steering baselines across three tasks, achieving stronger behavioral alignment with minimal impact on fluency or coherence.
    \item We show that meaningful control can be attained with only a small subset of latent dimensions, enhancing both steering interpretability and intervention efficiency.
\end{itemize}

\section{Preliminaries}

\subsection{Latent Steering in Language Models}
\label{preli}
Steering is a technique for controlling the output of LLMs via small interventions in their internal representations. Let $x$ be an input sequence in LLM and $h(x) \in \mathbb{R}^d$ denote the activation of $x$ at a chosen layer (e.g., the residual stream). Steering can modify $h(x)$ with an additive perturbation $v \in \mathbb{R}^d$ ($\lambda \in \mathbb{R}$ is a scaling coefficient) like ~\autoref{eq1}:
\begin{equation}
\label{eq1}
h'(x) = h(x) + \lambda v,
\end{equation}
The modified representation $h'(x)$ is then fed into the subsequent layers of the language model, thereby influencing the final output generation.

Prior work proposed various ways to construct the additive perturbation vector $v$, including mean difference vectors between contrasting classes~\citep{dathathri2020plug}, and PCA~\citep{kleindessner2023efficient}.
These approaches aim to identify semantically meaningful directions in the latent space, such that steering along these directions enables controlled manipulation of the LLM's behavior during generation.

\begin{figure*}[ht]
    \centering
    \includegraphics[width=0.98\textwidth]{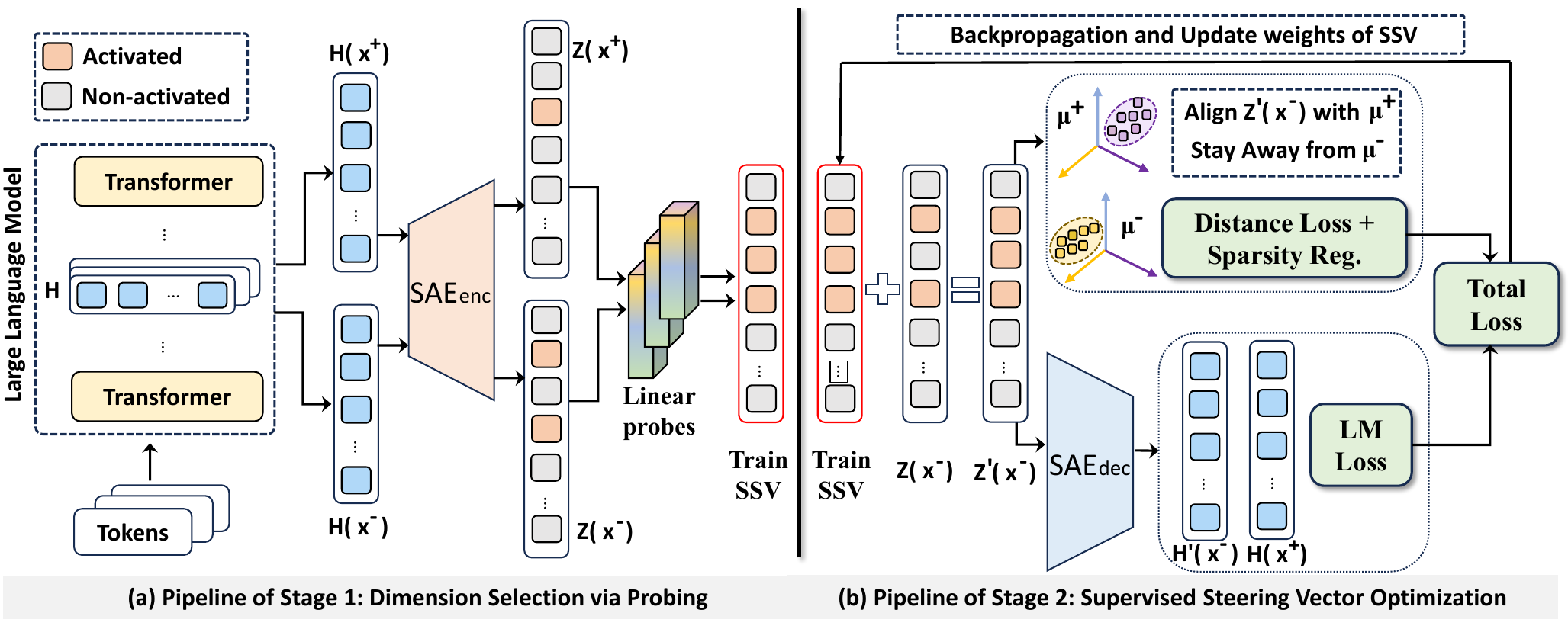}
    \caption{\textbf{Overview of the SAE-SSV framework.} It encodes model activations into a sparse latent space, selects task-relevant dimensions via linear probes, and optimizes steering vectors with combined losses to ensure effective control while maintaining generation quality.}
    \label{fig:method}
\end{figure*}

\subsection{SAEs for Representation Analysis}

To enable structured and interpretable analysis of internal model representations, Sparse autoencoders (SAEs) have been introduced to transform dense activations into sparse latent codes~\cite{huben2024sparse}. An SAE consists of an encoder $f_{\text{enc}}$ and decoder $f_{\text{dec}}$, trained to minimize \autoref{eq:sae_loss}:
\begin{align}
&z = f_{\text{enc}}(h), \quad \hat{h} = f_{\text{dec}}(z), \label{eq:sae_encdec} \\
&\mathcal{L}_{\text{SAE}} = \| h - \hat{h} \|_2^2 + \beta \| z \|_1, \label{eq:sae_loss}
\end{align}
where $h \in \mathbb{R}^m$ is the original activation vector of input, $z \in \mathbb{R}^{d_{\text{sae}}}$ is the sparse space, where typically $d_{\text{sae}} \gg m$ to allow for disentangled features. $\beta$ controls the sparsity, the $\ell_1$ penalty encourages each input to activate only a small number of latent dimensions, facilitating interpretability and localization of concepts \citep{bricken2023monosemanticity}.

\section{SAE-SSV Framework}
\label{sec:framework}
Our objective is to reliably steer the LLM’s output toward specific behavioral targets, such as producing text with a particular emotion. To achieve this, we propose the SAE-SSV framework (see Figure~\ref{fig:method}).
We first train multiple linear classifiers on labeled examples in the SAE space to identify a task-specific subspace relevant to the steering task (as ~\autoref{sec:dimension-selection}). We then learn a sparse steering vector within this subspace, optimized to shift representations toward the target class while preserving generation quality (as ~\autoref{sec:supervised-steering}).

\subsection{Dimension Selection via Probing}
\label{sec:dimension-selection}

\noindent\textbf{Coarse-Grained Feature Selection.}\,
We begin by identifying \textit{which dimensions in the SAE space are informative} for the steering task. Given a labeled dataset $D = \{(x_i, y_i)\}_{i=1}^N$, where $y_i \in \{0, 1\}$ denotes a binary attribute (e.g., negative vs.\ positive sentiment), we process each input $x_i$ through a frozen pretrained LLM and extract residual stream activations $h_i$ at a target layer as described in \autoref{preli}.

These activations are passed through a pretrained SAE encoder $f_{\text{enc}}$ to obtain sparse latent representations $z_i = f_{\text{enc}}(h_i)$. To identify task-relevant features, we compute the F-statistic~\cite{jain2002feature} for each latent dimension $t$:
\begin{equation}
S_t = \frac{\text{Between-group variance}}{\text{Within-group variance}},
\end{equation}
where the numerator quantifies how distinct the class means are and the denominator captures within-class dispersion. We rank all dimensions by $S_t$ and select the top-$k$ to form the steering subspace $I \subset [1, d_{\text{sae}}]$, where $d_{\text{sae}}$ is the dimensionality of the full SAE space. Representations restricted to $I$ are standardized and used to train a linear classifier to distinguish between the two classes. The classifier is optimized using the standard cross-entropy loss:
\begin{equation}
\label{eq:ce-loss}
\mathcal{L}_{\text{clf}} = \mathbb{E}_{(z, y) \sim D} \left[ -\log \frac{\exp(w_y^\top z)}{\sum_{y'} \exp(w_{y'}^\top z)} \right]
\end{equation}
where $w_y$ denotes the weight vector for class $y$. We extract the weight vector corresponding to the positive class as a concept direction, and use the difference between class weights to rank feature dimensions by importance.

\noindent\textbf{Fine-Grained Feature Selection.}\,
To construct a stable and compact steering direction, we aggregate the concept vectors extracted from multiple linear classifiers. Specifically, we train $M$ classifiers on independently sampled subsets of the data, using only the $k$ dimensions selected in the previous step. From each classifier, we extract the weight vector associated with the positive class label, denoted $w_1^{(j)}$ for the $j$-th classifier. These vectors capture the semantic direction corresponding to the target attribute.We compute the average of these vectors to obtain a unified direction:
\begin{equation}
\label{eq:avg-concept}
v_{\text{avg}} = \frac{1}{M} \sum_{j=1}^M w_1^{(j)}.
\end{equation}
This averaged vector serves as a representative semantic direction that consolidates information across multiple probing classifiers.

To further reduce dimensionality, we sort the coordinates of $v_{\text{avg}}$ by absolute magnitude and construct truncated vectors $v^{(d)}$ by retaining only the top-$d$ components and zeroing out the rest. For each $d$, we project test samples onto $v^{(d)}$ and compute their cosine similarity with the direction. Let $\bar{c}_1$ and $\bar{c}_0$ denote the average cosine similarity for positive and negative examples, respectively. We define the separation score as
\begin{equation}
\label{eq:separation-score}
s^{(d)} = \bar{c}_1 - \bar{c}_0.
\end{equation}
We select the smallest $d$ that maximizes $s^{(d)}$ and denote it as $d_{\text{steer}}$, which represents the final number of active dimensions used for steering.

\subsection{Supervised Steering Vector Optimization}
\label{sec:supervised-steering}

We construct and optimize a steering vector \(v \in \mathbb{R}^{d_{\text{sae}}}\) that is constrained to be nonzero only in the \(d_{\text{steer}}\) most informative dimensions, as identified in Section~\ref{sec:dimension-selection}. All remaining coordinates of \(v\) are fixed to zero, leaving only \(d_{\text{steer}}\) nonzero entries corresponding to the selected dimensions.

We initialize $v$ using the difference between class centroids in the SAE space:
\begin{equation}
v_{\text{init}} = \mu^+ - \mu^-,
\end{equation}
where $\mu^+$ and $\mu^-$ denote the average SAE representations of positive and negative examples, respectively. We then zero out all components of $v_{\text{init}}$ outside $I$, retain the top-$d_{\text{steer}}$ coordinates by magnitude, and normalize the resulting vector.

To optimize $v$, we construct training pairs $(x^+, x^-)$ of positive and negative examples. For each negative input $x^-$, we extract its SAE latent representation $z = f_{\text{enc}}(h(x^-))$, apply the steering vector to obtain $z' = z + v$, decode $z'$ back to the residual stream via $\hat{h} = f_{\text{dec}}(z')$, and reinsert it into the LLM to generate steered output.

The steering vector is optimized to satisfy three objectives: (1) align $z'$ with the positive class center while pushing it away from the negative center, (2) preserve the fluency and coherence of the generated text, and (3) maintain sparsity over the active dimensions. The total loss is given by:
\begin{equation} \label{eq:steer_split}
\begin{split}
L_{\text{steer}} ={}& \|z' - \mu^+\|_2^2 - \|z' - \mu^-\|_2^2 \\
                   & + L_{\text{LM}} + \beta \|v_I\|_1,
\end{split}
\end{equation}
where $L_{\text{LM}}$ is a language modeling loss that penalizes degraded generation quality by computing the cross-entropy of the positive target sequence $x^+$ conditioned on the steered hidden state of the negative input $x^-$, and $\|v_I\|_1$ encourages sparsity within the steering subspace.

\begin{table*}[tp]
\centering
\setlength{\tabcolsep}{3pt}
\caption{Comparison of Steering Methods Across All Models and Tasks (Sentiment, Politics Polarity, Truthfulness)}
\label{tab:all_models_comparison_fullwidth}
\resizebox{1.0\textwidth}{!}{%
\begin{tabular}{@{}l l 
  S[table-format=2.1] S[table-format=-1.2] S[table-format=-1.2] 
  S[table-format=2.1] S[table-format=-1.2] S[table-format=-1.2] 
  S[table-format=2.1] S[table-format=-1.2] S[table-format=-1.2]@{}}
\toprule
\multirow{2}{*}{Model} & \multirow{2}{*}{Method} 
  & \multicolumn{3}{c}{Sentiment} 
  & \multicolumn{3}{c}{Politics Polarity} 
  & \multicolumn{3}{c}{Truthfulness} \\
\cmidrule(lr){3-5} \cmidrule(lr){6-8} \cmidrule(lr){9-11}
& & {SR (\%)}$\uparrow$ & {$\Delta$MTLD}$\uparrow$ & {$\Delta$Entropy$\uparrow$} 
  & {SR (\%)}$\uparrow$ & {$\Delta$MTLD}$\uparrow$ & {$\Delta$Entropy$\uparrow$} 
  & {SR (\%)}$\uparrow$ & {$\Delta$MTLD}$\uparrow$ & {$\Delta$Entropy}$\uparrow$ \\
\midrule
\multirow{5}{*}{Llama3.1-8B}
  & CAA~\cite{rimsky2024steering} & 45.6 & -0.35 & -0.19 & 43.7 & -0.27 & -0.16 & 28.7 & -0.72 & -1.10 \\
  & RePe~\cite{zou2023representation} & 24.7 & -0.21 & -0.13 & 26.2 & -0.22 & -0.15 & 16.2 & -0.57 & -0.47 \\
  & Top PC~\cite{im2025unified} & 28.4 & -0.25 & -0.14 & 24.3 & -0.18 & -0.11 & 14.9 & -0.61 & -0.66 \\
  & ITI~\cite{li2024inference} & 41.1 & -0.31 & -0.27 & 45.2 & -0.34 & -0.29 & 31.2 & -0.81 & -0.89 \\
  & \cem{\textbf{SAE-SSV (Ours)}} & \cem\textbf{63.2} & \cem\textbf{0.09} & \cem\textbf{-0.07} & \cem\textbf{60.5} & \cem\textbf{0.11} & \cem\textbf{-0.04} & \cem\textbf{34.1} & \cem\textbf{-0.31} & \cem\textbf{-0.24} \\
\midrule
\multirow{5}{*}{Gemma2-2B}
  & CAA~\cite{rimsky2024steering} & 39.6 & -0.32 & -0.28 & 45.4 & -0.38 & -0.33 & 24.6 & -0.71 & -1.05 \\
  & RePe~\cite{zou2023representation} & 27.2 & -0.24 & -0.20 & 36.6 & -0.26 & -0.21 & 11.6 & -0.42 & -0.37 \\
  & Top PC~\cite{im2025unified} & 23.8 & -0.17 & -0.09 & 35.0 & -0.22 & -0.17 & 12.2 & -0.46 & -0.40 \\
  & ITI~\cite{li2024inference} & 41.2 & -0.30 & -0.27 & 42.1 & -0.33 & -0.32 & 22.3 & -0.74 & -1.12 \\
  & \cem{\textbf{SAE-SSV (Ours)}} & \cem\textbf{52.8} & \cem\textbf{0.08} & \cem\textbf{-0.08} & \cem\textbf{61.3} & \cem\textbf{0.10} & \cem\textbf{-0.04} & \cem\textbf{31.7} & \cem\textbf{-0.37} & \cem\textbf{-0.23} \\
\midrule
\multirow{5}{*}{Gemma2-9B}
  & CAA~\cite{rimsky2024steering} & 42.3 & -0.42 & -0.37 & 39.3 & -0.27 & -0.22 & 19.8 & -0.75 & -1.10 \\
  & RePe~\cite{zou2023representation} & 19.7 & -0.27 & -0.22 & 22.4 & -0.16 & -0.19 & 9.2 & -0.51 & -0.48 \\
  & Top PC~\cite{im2025unified} & 21.4 & -0.31 & -0.25 & 29.1 & -0.21 & -0.18 & 10.6 & -0.66 & -0.70 \\
  & ITI~\cite{li2024inference} & 41.2 & -0.33 & -0.29 & 33.8 & -0.31 & -0.27 & 21.4 & -0.70 & -0.97 \\
  & \cem{\textbf{SAE-SSV (Ours)}} & \cem\textbf{48.5} & \cem\textbf{0.09} & \cem\textbf{-0.11} & \cem\textbf{55.0} & \cem\textbf{0.07} & \cem\textbf{-0.12} & \cem\textbf{27.2} & \cem\textbf{-0.39} & \cem\textbf{-0.26} \\
\bottomrule
\end{tabular}%
}
\end{table*}

\begin{figure*}[t] 
    \centering 

    \begin{subfigure}[b]{0.48\textwidth} 
        \centering
        \includegraphics[width=\linewidth]{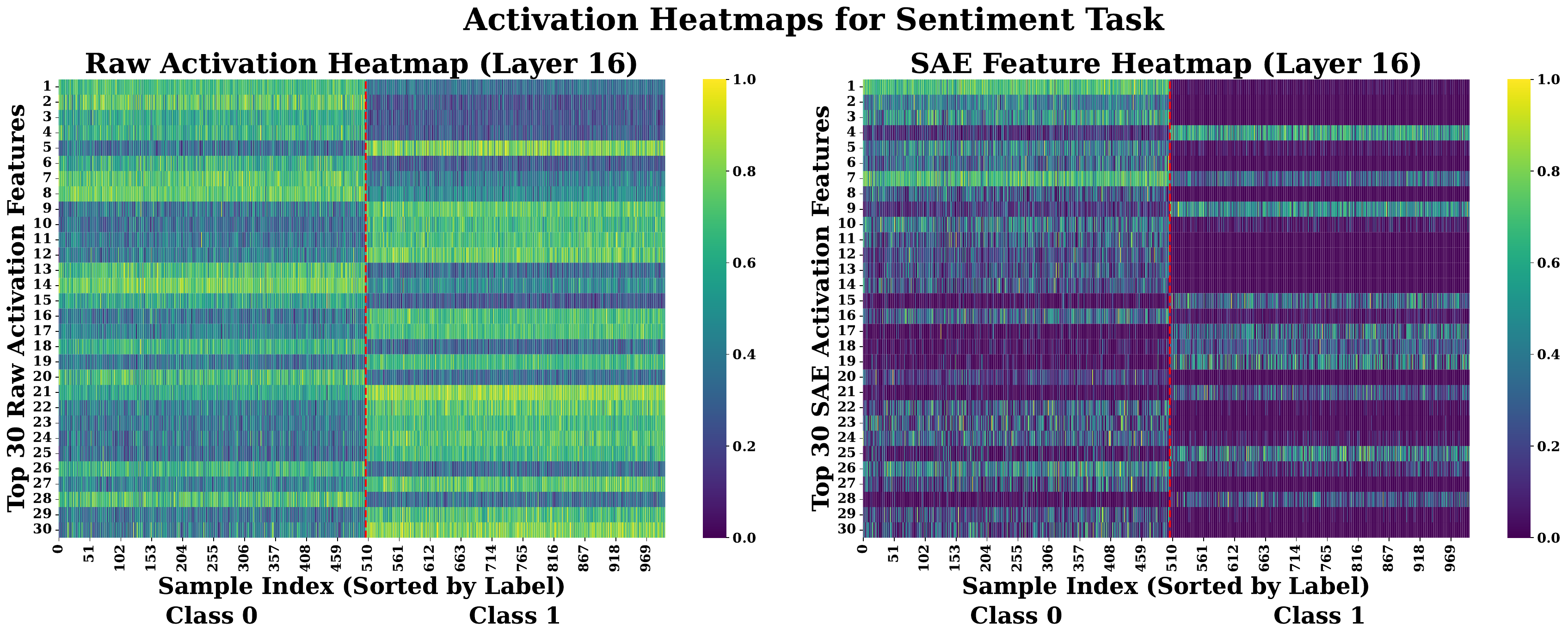}
        \caption{Sentiment Task (LLaMA3.1-8b, Layer 16)}
        \label{fig:heatmapsub1} 
    \end{subfigure}
    \hfill 
    \begin{subfigure}[b]{0.48\textwidth} 
        \centering
        \includegraphics[width=\linewidth]{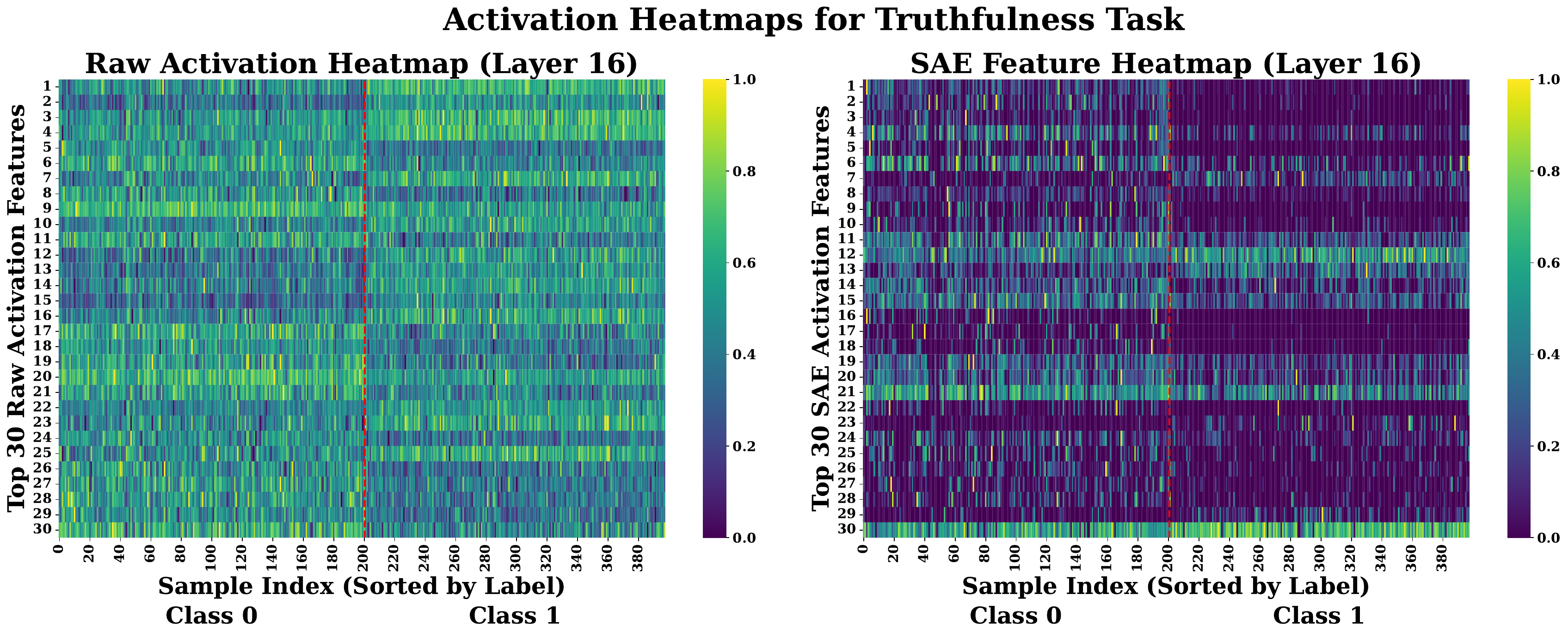}
        \caption{Truthfulness Task (LLaMA3.1-8b, Layer 16)}
        \label{fig:heatmapsub2} 
    \end{subfigure}

    \caption{Activation heatmaps of the top-30 dimensions for each task. (a) Sentiment task. (b) Truthfulness task. Each panel compares class-wise activation patterns in the raw residual space and SAE space.}
    \label{fig:heatmap} 

\end{figure*}

\section{Experiments}
In this section, we  evaluate the effectiveness of SAE-SSV by answering the following research questions (RQs):
\vspace{-5pt}
\begin{itemize}[leftmargin=*]\setlength\itemsep{-0.3em}
\item RQ1: How is the performance of SAE-SSV compared to baselines? (Section 4.2) 
   \item RQ2: Can we identify a minimal and interpretable subspace within the SAE latent space that is sufficient for steering model behavior? (Section 4.3)
   
   \item RQ3: Can steering in a structured subspace improve attribute alignment while minimizing output degradation? (Section 4.4)
   
   \item RQ4: Can SAE-SSV generalize across datasets within the same task domain? (Section 4.5)
\end{itemize}

\subsection{Experimental Setup}

\noindent\textbf{Models.}\, We conduct experiments on three open-source base models: Gemma-2-2b, Gemma-2-9b~\cite{team2024gemma}, and LLaMA3.1-8B~\cite{grattafiori2024llama}. For sparse autoencoders, we use pre-trained SAEs from the Gemma Scope~\cite{lieberum2024gemma} and LLaMA Scope~\cite{he2024llama} repositories to extract semantic subspaces for steering.

\noindent\textbf{Datasets.} \, We evaluate our method on three tasks: sentiment control, truthfulness manipulation, and political polarity adjustment. The truthfulness and political polarity datasets are adopted from~\cite{fulay2024relationship}, namely the \emph{TruthGen} dataset of paired factual and counterfactual statements, and the \emph{TwinViews-13k} dataset of ideologically matched political pairs. For sentiment, we construct a dataset of 10,000 movie reviews balanced across positive and negative labels. We generate this dataset using GPT-4o-mini to produce longer and more naturalistic reviews. 

\vspace{5pt}
\noindent\textbf{Baseline Methods.} 
We compare our SAE-SSV method against four widely used steering baselines:

\begin{itemize}[leftmargin=*]\setlength\itemsep{-0.3em}
\item \textit{Concept Activation Addition (CAA)}~\cite{rimsky2024steering}: Adds the mean activation difference between positive and negative examples during inference to steer model outputs.
\item \textit{Representation Perturbation (RePe)}~\cite{zou2023representation}: Perturbs activations along principal components of class-conditional differences.
\item \textit{Top PC}~\cite{im2025unified}: Projects activations onto the first principal component of the embedding space, capturing the direction of maximal variance.
\item \textit{Inference-Time Intervention (ITI)}~\cite{li2024inference}: Shifts attention head activations during inference along truth-related directions found via linear probing.
\end{itemize}

\vspace{5pt}
\noindent\textbf{Evaluation Metrics.} We employ metrics to evaluate steering effectiveness and generation quality:
\begin{itemize}[leftmargin=*]\setlength\itemsep{-0.3em}
\item \textit{Steering Success Rate (SR)}: The percentage of generated outputs that successfully exhibit the target attribute. We use GPT-4o-mini as an automatic judge to assess whether the generated text reflects the intended attribute. Formally, $\text{SR} = \frac{N_{\text{success}}}{N_{\text{total}}} \times 100\%$, where $N_{\text{success}}$ is the number of generations judged as exhibiting the target attribute. Specific prompting details for the judge are provided in Appendix~\ref{appendix:evaluation}.

\item \textit{Lexical Diversity (MTLD)}: Measures vocabulary richness based on the average length of text segments with stable type-token ratio (TTR). We report $\Delta\text{MTLD}$ relative to unsteered outputs to assess changes in lexical diversity.

\item \textit{Entropy}: Measures the unpredictability of token distributions using Shannon entropy. Lower values indicate higher repetition. We report $\Delta\text{Entropy}$ relative to unsteered outputs. Formally, $H = -\sum_{i} p(x_i) \log p(x_i)$, where $p(x_i)$ is the probability of token $x_i$.
\end{itemize}


\begin{figure*}[t] 
    \centering 

    \begin{subfigure}[b]{0.44\textwidth} 
        \centering
        \includegraphics[width=\linewidth]{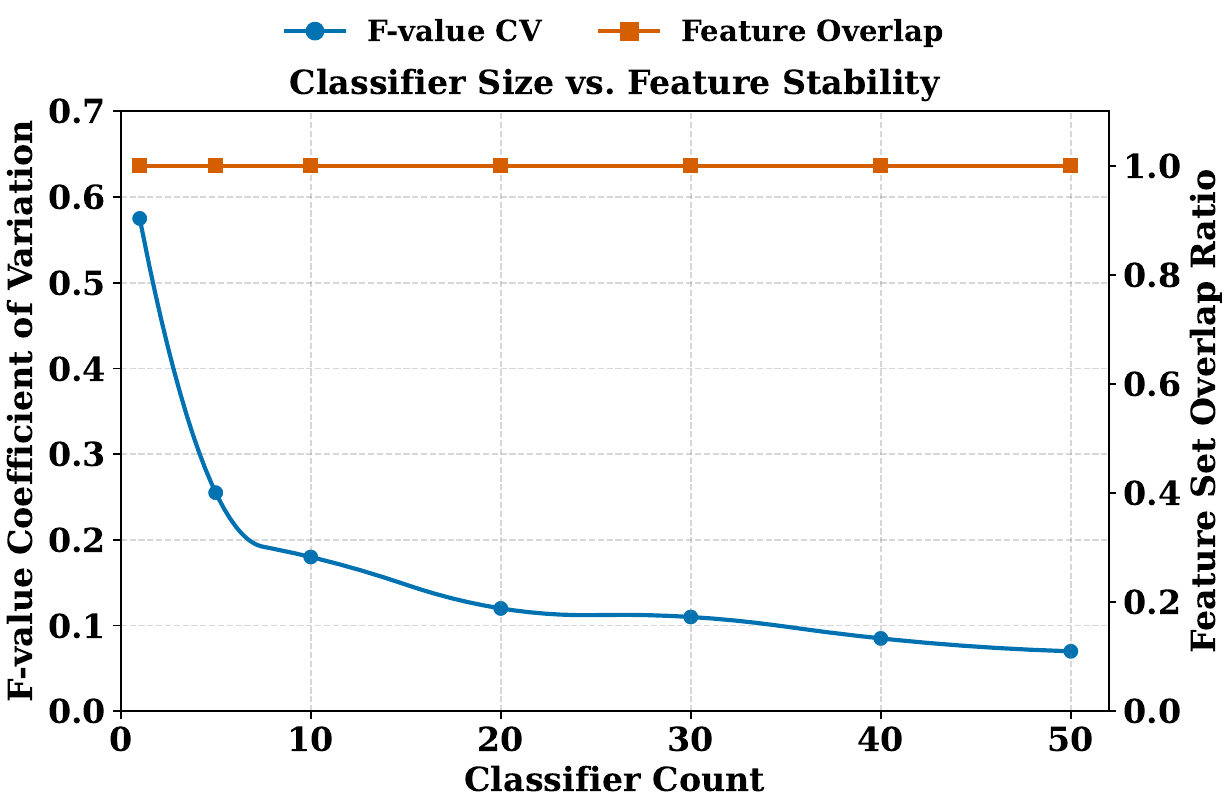}
        \caption{Feature Selection Stability} 
        \label{fig:dimsub1} 
    \end{subfigure}
    \hfill 
    \begin{subfigure}[b]{0.48\textwidth} 
        \centering
        \includegraphics[width=\linewidth]{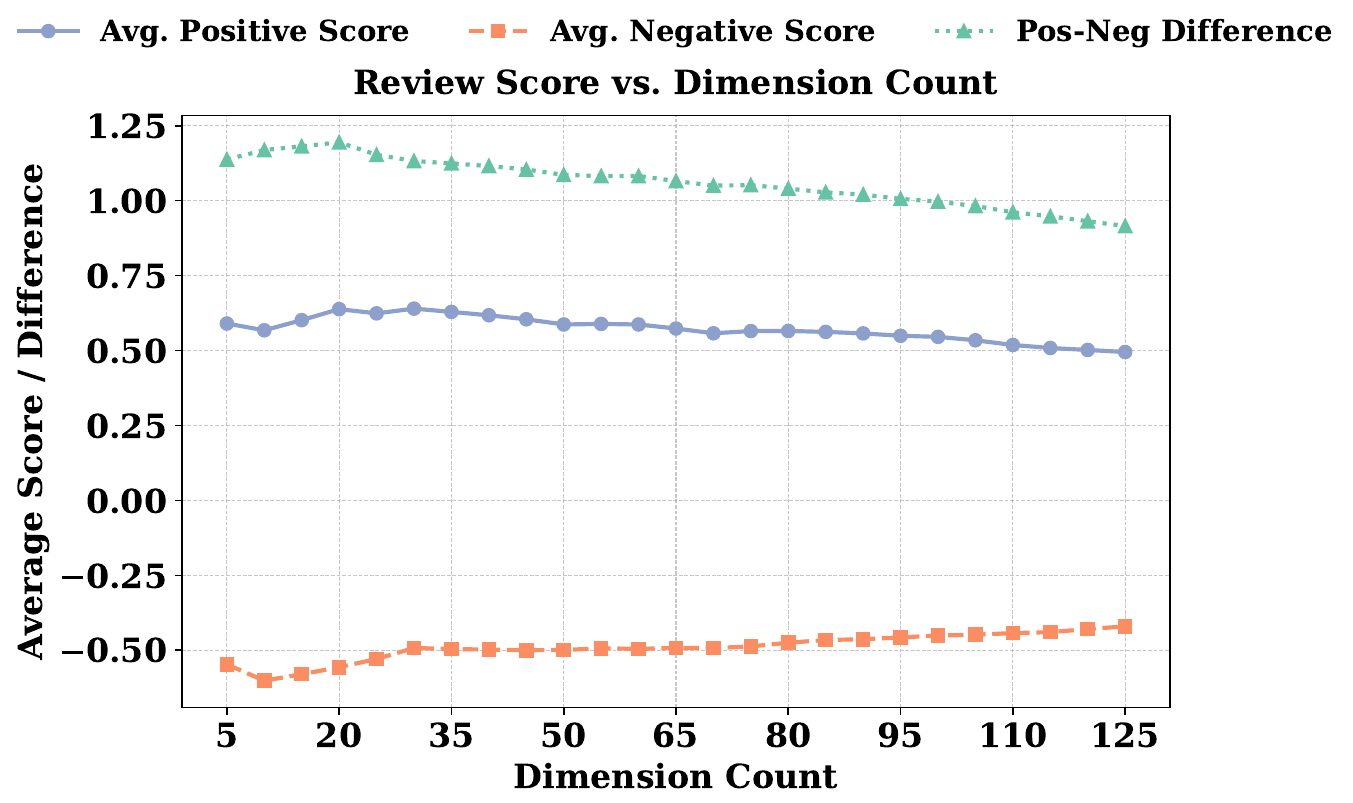}
        \caption{Separability vs. Dimension Count} 
        \label{fig:dimsub2} 
    \end{subfigure}

    \caption{(a) shows how the number of linear classifiers affects feature selection stability. (b) shows that a small number of top SAE dimensions enable clear class separation.}
    \label{fig:dimension_analysis} 

\end{figure*}

\noindent\textbf{Implementation Details.} 
We report main results using 16K-dimensional SAE models for both Gemma-2-2b and Gemma-2-9b models, and a 32K-dimensional SAE for LLaMA3.1-8b model. 
Following our methodology in Section~\ref{sec:framework}, we adopt a two-stage steering pipeline. 
In Stage 1, we train $M = 50$ linear probes per task to ensure stability in feature selection. 
We set the number of selected SAE dimensions to $k = 128$ to ensure sufficient subspace coverage for semantic manipulation. 
In Stage 2, the steering vector is optimized using a contrastive objective that combines distance loss, language modeling loss, and $L_1$ regularization, with coefficients $\lambda_{\mathrm{dist}} = 1.0$, $\lambda_{\mathrm{lm}} = 0.5$, and $\lambda_{\mathrm{reg}} = 0.01$, respectively. 
Optimization is performed for 100 iterations with a learning rate of 0.05 and a batch size of 64. 
During inference, we apply the steering vector at each decoding step with scaling factors ranging from 1.0 to 10.0 to explore the trade-off between steering strength and output quality. For each model and task, we apply interventions at empirically selected layers: 
LLaMA3.1-8B (layer 16 for all tasks), 
Gemma2-2B (sentiment: 13, truthfulness: 16, politics: 15), 
and Gemma2-9B (sentiment: 20, truthfulness: 26, politics: 20).
All experiments are conducted on a single NVIDIA A100 GPU.

\subsection{Comparison with Baseline  Methods}

For each task, we steer in a fixed semantic direction: from negative to positive sentiment, from left-leaning to right-leaning political views, and from factual to hallucinated content. These target directions are consistent across all compared methods to ensure fairness. 
Table~\ref{tab:all_models_comparison_fullwidth} presents steering comparison across the three tasks. We have the following observations.

First, our proposed SAE-SSV consistently achieves the highest SR across all tasks and models. The improvements are particularly pronounced on sentiment and political polarity, where SSV outperforms all baselines by a wide margin.
Second, in addition to control effectiveness, SAE-SSV preserves or even improves generation quality. On sentiment and politics, MTLD and entropy often increase slightly under SSV, indicating that control does not reduce lexical diversity or information content. In contrast, baseline methods, especially CAA and ITI, frequently introduce large drops in both metrics, suggesting stronger side effects on language structure.
Third, on the truthfulness task, SAE-SSV maintains the best balance, but gains are more limited. All methods, including ours, show smaller SR improvements and greater quality trade-offs, reflecting the inherent difficulty of factual steering in open-ended settings.

\subsection{Identifying a Minimal Steering Subspace}
\label{sec:RQ1}
We investigate whether the model's internal representations contain a sparse and semantically aligned subspace that supports effective steering. 

\vspace{3pt}\noindent
\textbf{Subspace Concept Separability Analysis.}
Figure~\ref{fig:heatmap} compares average activation patterns in both the residual stream and the SAE-encoded space, using positive and negative samples. We visualize the top 20 most active dimensions in each space. We have two key observations:
\begin{itemize}[leftmargin=*]\setlength\itemsep{-0.3em}
\item In the residual space, activations are distributed without clear class-specific structure. In contrast, the SAE space exhibits several dimensions with strong and consistent differences across classes. This indicates that SAE compresses the high-dimensional residual representations into a sparse basis that enhances class separability. It suggests that \emph{the SAE latent space is a promising domain for constructing effective steering vectors}.

\item The SAE heatmaps also reveal task-specific characteristics. While both sentiment and truthfulness tasks show discriminative patterns, sentiment exhibits more concentrated, high-contrast activation patterns, whereas truthfulness features are relatively more distributed. This structural difference in the representation space aligns with the performance patterns in Table~\ref{tab:all_models_comparison_fullwidth}, where our method achieves higher success rates on sentiment and politics polarity tasks (SR = 48.5-63.2\%) compared to truthfulness (SR = 27.2-34.1\%). Also, even on the more challenging truthfulness task, our method still substantially outperforms all baselines, demonstrating that \emph{our sparse subspace approach effectively captures key features across different types of tasks}.
\end{itemize}

\begin{table*}[tb]
\centering
\caption{Top-10 SAE features used in the SSV for the sentiment task on \texttt{LLaMA-3.1-8B}. Feature explanations are retrieved from Neuronpedia~\cite{neuronpedia}, and the value column indicates the weights learned during SSV training.}
\label{tab:llama_sentiment}
\setlength{\tabcolsep}{4pt} 
\renewcommand{\arraystretch}{1.1} 
\small 
\begin{tabular}{@{}r p{0.56\textwidth} c S[table-format=2.2]@{}}
\toprule
\textbf{Rank} & \textbf{Explanation of Feature} & \textbf{SAE Feature \#} & \textbf{Value} \\
\midrule
1  & Negative sentiments towards characters in movies                       & 2305  & -6.76 \\
2  & Negative sentiments and criticisms related to performance or quality  & 14086 & -3.24 \\
3  & Phrases related to notable achievements in the entertainment industry & 12322 &  2.79 \\
4  & Punctuation and symbols indicative of structural elements in text     & 20767 &  2.32 \\
5  & Mentions of achievements and recognition in a professional context    & 28857 &  1.99 \\
6  & Phrases and terminology related to legal injunctions and restrictions & 2268  & -1.89 \\
7  & Components related to specific abilities or skills in performance     & 29039 & -1.86 \\
8  & References to historical or legendary figures and events              & 28858 & -1.68 \\
9  & Phrases indicating misinformation, contradictions, or inaccuracies    & 14391 & -1.46 \\
10 & Questions and expressions of disappointment or concern                & 13758 & -1.43 \\
\bottomrule
\end{tabular}
\label{tab:top-10-feature-main}
\end{table*}

\vspace{3pt}\noindent
\textbf{Feature Selection Stability Analysis.}
We vary the number of linear classifiers $M$ used to rank important SAE dimensions. Each classifier is trained on a random subset of labeled data, and we compute the average importance scores across all $M$ runs. Figure~\ref{fig:dimsub1} demonstrates that despite variations in their relative rankings, the set of top-128 dimensions selected from the 16K-dimensional SAE space remains perfectly consistent across different ensemble sizes ($M=1$ to $M=50$). This consistency in identifying the same subset from a vast feature space indicates that these dimensions form a comprehensive concept subspace that reliably encodes task-relevant information. The coefficient of variation of feature importance scores decreases as $M$ increases, providing more stable estimates of each dimension's relative contribution.

\vspace{3pt}\noindent
\textbf{Selected Dimension Discriminability Analysis.}

\begin{figure}[tbp]
  \centering
  \includegraphics[width=\linewidth]{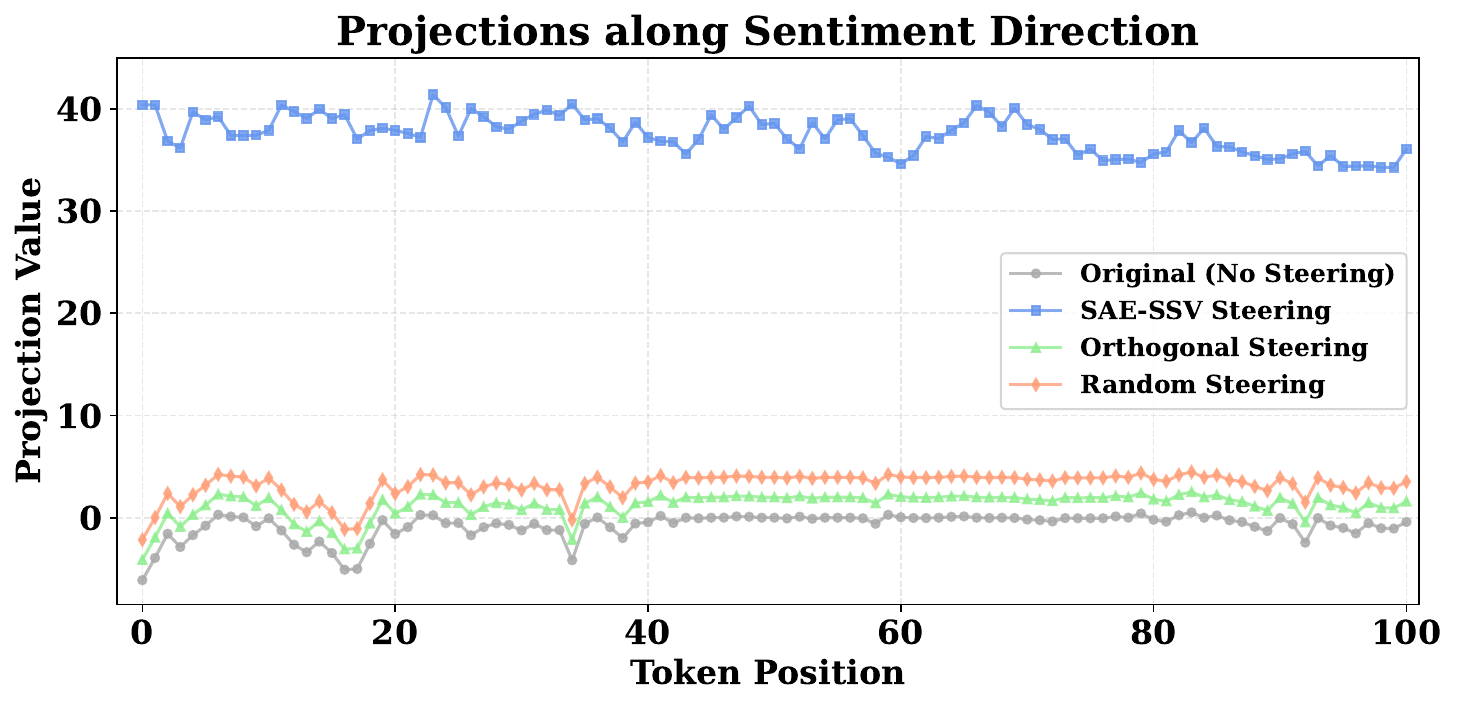}
  \caption{Average projection values of token activations along four directions: no steering (gray), SAE-SSV (blue), orthogonal (green), and random (orange). Computed over successfully steered samples. SAE-SSV induces a consistent and sustained directional shift, while other directions show minimal change.}
  \label{fig:sep_projections}
\end{figure}
\vspace{-2pt}

In Figure~\ref{fig:dimsub2}, we incrementally select the top-$k$ ranked dimensions from our identified 128-dimensional subspace and measure class separability by calculating the difference between mean projection scores of positive and negative samples. The results demonstrate that even a small number of the highest-ranked dimensions achieves substantial class separation, with diminishing returns as more dimensions are added. This suggests that within our already focused 128-dimensional concept space, an even smaller subset of dimensions carries the most significant task-relevant information. This finding supports our approach of extremely targeted steering interventions, where \emph{modifications to just a small fraction of the SAE space can effectively influence specific attributes} while maintaining computational efficiency. We provide the sets of SAE features used for constructing SSVs in Table~\ref{tab:top-10-feature-main} and more analysis in Appendix~\ref{appendix:SAE features}.

\subsection{Mitigating Output Degradation}
We evaluate whether SAE-SSV can achieve strong steering while minimizing generation quality degradation, a common side effect of intervention. 

\vspace{3pt}\noindent
\textbf{Measuring Output Degradation Quality.}
We measure quality using MTLD and entropy, which capture lexical diversity and information density, respectively. As shown in Table~\ref{tab:all_models_comparison_fullwidth}, SAE-SSV consistently improves or preserves these metrics on the sentiment and politics tasks. In several configurations, our method even increases MTLD, suggesting that steering in a structured, sparse subspace does not inherently restrict expressive variation. 
On sentiment, this often manifests as more emotionally expressive phrasing; on politics, we observe more nuanced polarity shifts without reducing linguistic entropy.
Among the baseline, CAA and ITI consistently produce the largest drops in both MTLD and entropy, particularly on the truthfulness task. 

\vspace{3pt}\noindent
\textbf{Why SAE-SSV can Preserve Quality?}
To better understand this question, we visualize the token-wise projection of hidden activations along different directions. Figure~\ref{fig:sep_projections} compares generation with no steering, SSV steering, orthogonal direction, and random direction. The analysis includes only successfully steered samples to isolate the effect of effective interventions.
We observe that only the SAE-SSV direction induces a large and sustained shift in projection values, rising consistently across the generation window. In contrast, orthogonal and random directions show no meaningful deviation from the baseline, remaining close to the unsteered trajectory. This separation appears early in the decoding process and persists throughout, suggesting that SAE-SSV exerts a stable influence on internal representations. 
The consistency of this shift across all successful samples supports the conclusion that \emph{SAE-SSV modifies internal representation in a structured and consistent direction}.

\subsection{Generalizing SAE-SSV Across Tasks}

\begin{table}[t]
  \small
  \centering
  \caption{Generalization performance of SAE-SSV on unseen datasets using LLaMA3.1-8B. SR = steering success rate. Ret. = retained original attribute. Dis. = incoherent, repetitive, contradictory or task irrelevant output. All values are percentages.}
  \label{tab:ssv_generalization}
  \begin{tabular}{@{}lccc@{}}
    \toprule
    \textbf{Method} & \textbf{SR (\%)} $\uparrow$ & \textbf{Ret. (\%)} $\downarrow$ & \textbf{Dis. (\%)} $\downarrow$ \\
    \midrule
    \multicolumn{4}{@{}l}{\textit{Rotten Tomatoes}} \\
    Baseline & 20.2 & 63.1 & 16.7 \\
    SAE-SSV  & 37.8 & 33.5 & 28.7 \\
    \midrule
    \multicolumn{4}{@{}l}{\textit{TruthfulQA}} \\
    Baseline & 32.4 & 57.8 & 9.8 \\
    SAE-SSV  & 48.9 & 9.8 & 41.3 \\
    \bottomrule
  \end{tabular}
\end{table}

To evaluate the generalization capacity of our proposed SAE-SSV method, we apply steering vectors originally trained on one dataset to a different test set within the same task domain, without any retraining or supervision on the target samples.

\vspace{3pt}\noindent
\textbf{Experimental Setting.}
We test on two new datasets for open-ended generation: \textit{Rotten Tomatoes} for sentiment steering and \textit{TruthfulQA} for truthfulness steering. In the sentiment task, the steering direction targets \textit{positive sentiment}, while in the truthfulness task, the direction induces \textit{hallucinated} content.
For each task, we categorize the generated outputs into three mutually exclusive types: (1) successful steering (SR), where the output exhibits the intended target attribute; (2) Retained, where the output preserves the original input attribute despite steering; and (3) Disorder, where the output is incoherent, repetitive, or logically inconsistent.

\vspace{3pt}\noindent
\textbf{Result Analysis.}
As shown in Table~\ref{tab:ssv_generalization}, in the sentiment task, the baseline model mostly preserves the original negative tone, with an SR of 20.2\%. Applying the SAE-SSV vector raises SR to 37.8\%, demonstrating effective transfer of the emotional control signal. The Retained rate drops from 63.1\% to 33.5\%, suggesting that most outputs have been influenced by the steering. However, this comes with a trade-off, as the Disorder rate rises to 28.7\%, indicating more outputs falling into unusable forms.
On the truthfulness task, the baseline SR is 32.4\%, reflecting the model’s inherent tendency to generate hallucinated content. With SAE-SSV steering, SR increases to 48.9\%, and Retained drops sharply to 9.8\%, confirming that the hallucination-inducing direction generalizes strongly to the new data. 


\subsection{Ablation Study}
Table~\ref{tab:ablation_sentiment} examines the impact of two key components in our method: the supervised training of the steering vector and the inclusion of the LM loss. Removing either component leads to a clear drop in steering success. Notably, omitting the LM loss increases SR to 28.6\%, but also causes a substantial rise in output disorder (43.3\%), indicating unstable model behavior. In contrast, the full SAE-SSV achieves the highest SR (63.2\%) while maintaining low disorder (13.3\%), demonstrating the importance of subspace-constrained, supervised optimization. In addition, we study the effect of the scaling factor $\lambda$ used during inference. We observe that the steering strength measured qualitatively by semantic shift is approximately linear with respect to $\lambda$. However, developing a precise quantitative metric for steering intensity remains challenging. We provide representative examples illustrating this relationship in Appendix~\ref{appendix:factors}.

\begin{table}[t]
  \centering
  \small
  \caption{Ablation results for sentiment steering with LLaMA3.1-8B. 
We compare the full SAE-SSV with two ablated variants and the baseline.Evaluation metrics are identical to those in Table~\ref{tab:ssv_generalization}.}
  \label{tab:ablation_sentiment}
  \begin{tabular}{@{}lccc@{}}
    \toprule
    \textbf{Method} & \textbf{SR (\%)} $\uparrow$ & \textbf{Ret. (\%)} $\downarrow$ & \textbf{Dis. (\%)} $\downarrow$ \\
    \midrule
    Baseline & 12.3 & 79.2 & 8.5 \\
    SSV w/o train & 13.7 & 73.9 & 12.4 \\
    SSV w/o LM loss & 28.6 & 28.1 & 43.3 \\
    \midrule
    SSV & 63.2 & 23.5 & 13.3 \\
    \bottomrule
  \end{tabular}
\end{table}

\section{Related Work}

\paragraph{Language Model Representations.}
Studies of language model representations have established that many concepts exist as linear directions in activation space \cite{kim2018interpretability, jin2025massive}. These concept vectors can be derived through various methods, including probing classifiers \cite{belinkov2022probing, jin2025exploring}, mean difference calculations \cite{rimsky2024steering, zou2023representation}, mean centering \cite{jorgensen2024improving}, and Gaussian concept subspaces \cite{zhao2025beyond}. These approaches have successfully identified directions corresponding to high-level concepts such as honesty \cite{li2024inference}, truthfulness \cite{tigges2023linear}, harmfulness \cite{zou2023representation}, and sentiment \cite{zhao2025beyond}. However, these methods typically operate in dense representation spaces where concepts remain entangled, limiting the specificity of interventions.

\paragraph{Activation Steering.}
Activation steering has emerged as a powerful technique for influencing model behavior during inference without retraining. Early work such as Plug and Play Language Models \cite{dathathri2020plug} and representation engineering \cite{zou2023representation} established the feasibility of direct activation manipulation. Subsequent research demonstrated its effectiveness in improving truthfulness \cite{marks2023geometry, tigges2023linear}, enhancing safety \cite{arditi2024refusal, li2024inference}, mitigating biases \cite{jorgensen2024improving}, and controlling style \cite{wang2024steering}. More recent methods include CAA \cite{rimsky2024steering}, which uses contrastive activation addition, RePe \cite{kleindessner2023efficient}, which employs PCA-derived directions, and ITI \cite{li2024inference}, which iteratively trains steering vectors. Nevertheless, steering often faces a trade-off between control strength and generation quality in open-ended settings \cite{zhou2024evaluating}, in part because interventions in dense spaces can inadvertently entangle multiple concepts \cite{huben2024sparse}. Our work addresses this challenge by leveraging disentangled SAE features and supervised dimension selection to constrain steering to a task-specific subspace, enabling more targeted interventions with fewer side effects.

\paragraph{Sparse Autoencoders.}
Sparse autoencoders (SAEs) have been introduced to disentangle superimposed features through dictionary learning. By mapping activations into a higher-dimensional sparse space, SAEs yield more interpretable features \cite{bricken2023monosemanticity, huben2024sparse}. Variants include vanilla SAEs \cite{sharkey2022sae} and TopK SAEs \cite{gao_scaling_2024}, with pre-trained repositories such as Gemma Scope \cite{lieberum2024gemma} and Llama Scope \cite{he2024llama} enabling broader research. SAEs have been used to interpret model representations \cite{kissane2024interpreting}, to understand model capabilities \cite{ferrando2025do}, and to explore intersections with steering \cite{chalnev2024improving, he2025saif}. Applications include toxicity mitigation \cite{gallifant2025sparse} and safety alignment \cite{wu2025interpreting}, but the use of SAEs for controllable generation remains relatively limited. Our work extends this line by combining SAE-derived features with supervised optimization to construct effective steering vectors.

\section{Conclusions and Future Work}
In this paper, we introduced SAE-SSV, a framework that enables effective LLM steering by operating in sparse, task-specific subspaces. 
The key insight lies in constraining interventions to a small number of interpretable dimensions that capture task-relevant semantics, enabling more targeted control while preserving generation quality. Experiments across sentiment, truthfulness, and political polarity steering tasks with multiple LLMs demonstrate that SAE-SSV consistently outperforms existing methods by a substantial margin. Our cross dataset experiments reveal that SAE-SSV captures both semantic directions and stylistic patterns of training data, highlighting its potential as a more general steering mechanism. 
For our future work, we aim to achieve universal and style-invariant SSVs that generalize across datasets, tasks, and model families by curating diverse training corpora and developing objectives that explicitly encourage semantic steering while minimizing sensitivity to stylistic variation.

\section*{Limitations}
Our SAE-SSV approach has several limitations. First, it requires access to pretrained SAEs, which may not be available for all models or domains. Currently, we only evaluate using the Gemma and Llama model families.
Second, we evaluate LLMs with parameters at most of 9B. In future work, we plan to evaluate on larger LLMs with tens or hundreds of billions of parameters to better understand how our method scales with model size and complexity. Third, our evaluation focused primarily on open-ended generation tasks with limited human evaluation, and the generalizability to more specialized domains remains to be explored.

\section*{Acknowledgments}
Mengnan Du is supported by National Science Foundation (NSF) Grant \#2310261. The
views and conclusions in this paper are those of the authors and should not be interpreted as representing any funding agencies.

\bibliography{ref}

\clearpage
\appendix

\section{Case Study}
\label{appendix:case_study}

This appendix presents detailed case studies comparing model outputs under four steering conditions. Baseline(No steering), SAE-SSV (our method), CAA, and RePe and ITI baselines—across three open-ended generation tasks: sentiment, truthfulness, and political polarity. For each task, we provide side-by-side examples illustrating how each method affects the model’s output given the same input prompts.

Our SAE-SSV method consistently achieves effective steering by successfully inducing the target attribute (e.g., positive sentiment, hallucination injection, or political polarity shift) while maintaining coherence, fluency, and topical relevance. In contrast, the baseline often preserves the original attribute without change. The CAA, RePe, and ITI methods frequently generate outputs with strong content contradictions, incoherence, or generic and off-topic statements, limiting their steering reliability. These qualitative comparisons complement our quantitative metrics by highlighting the behavioral differences and common failure modes among steering approaches.

\begin{figure*}[htbp]
    \centering
    \includegraphics[width=\textwidth]{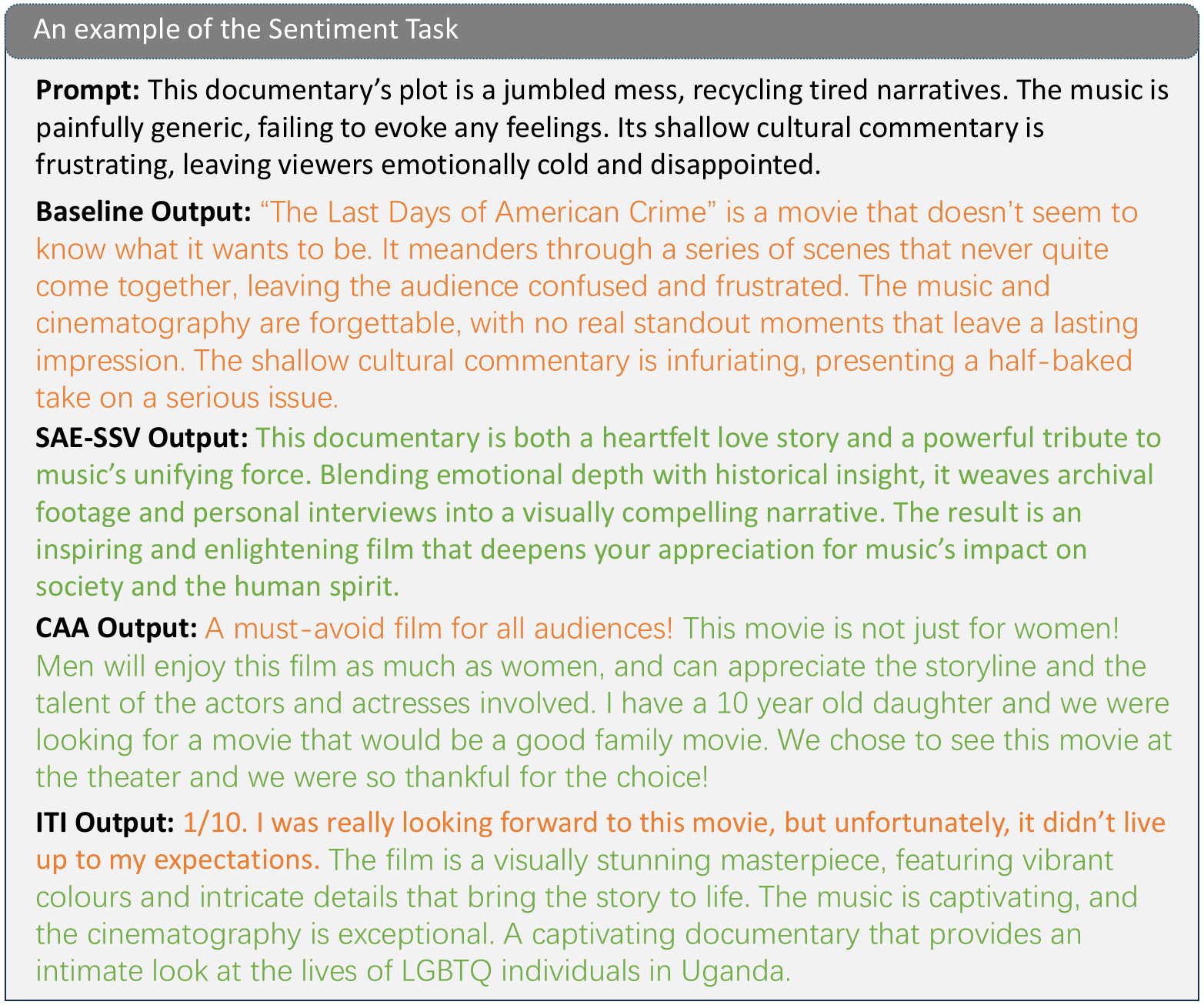}
    \caption{Case study on the sentiment steering task. The input prompts are negative movie reviews. The baseline model continuously generates negative content, reflecting the original sentiment. Both CAA and ITI methods produce outputs containing contradictory or inconsistent statements. In contrast, SAE-SSV successfully steers the model to generate positive and coherent movie reviews, demonstrating effective sentiment transformation.}
    \label{fig:sentiment_case}
\end{figure*}

\begin{figure*}[htb]
    \centering
    \includegraphics[width=\textwidth]{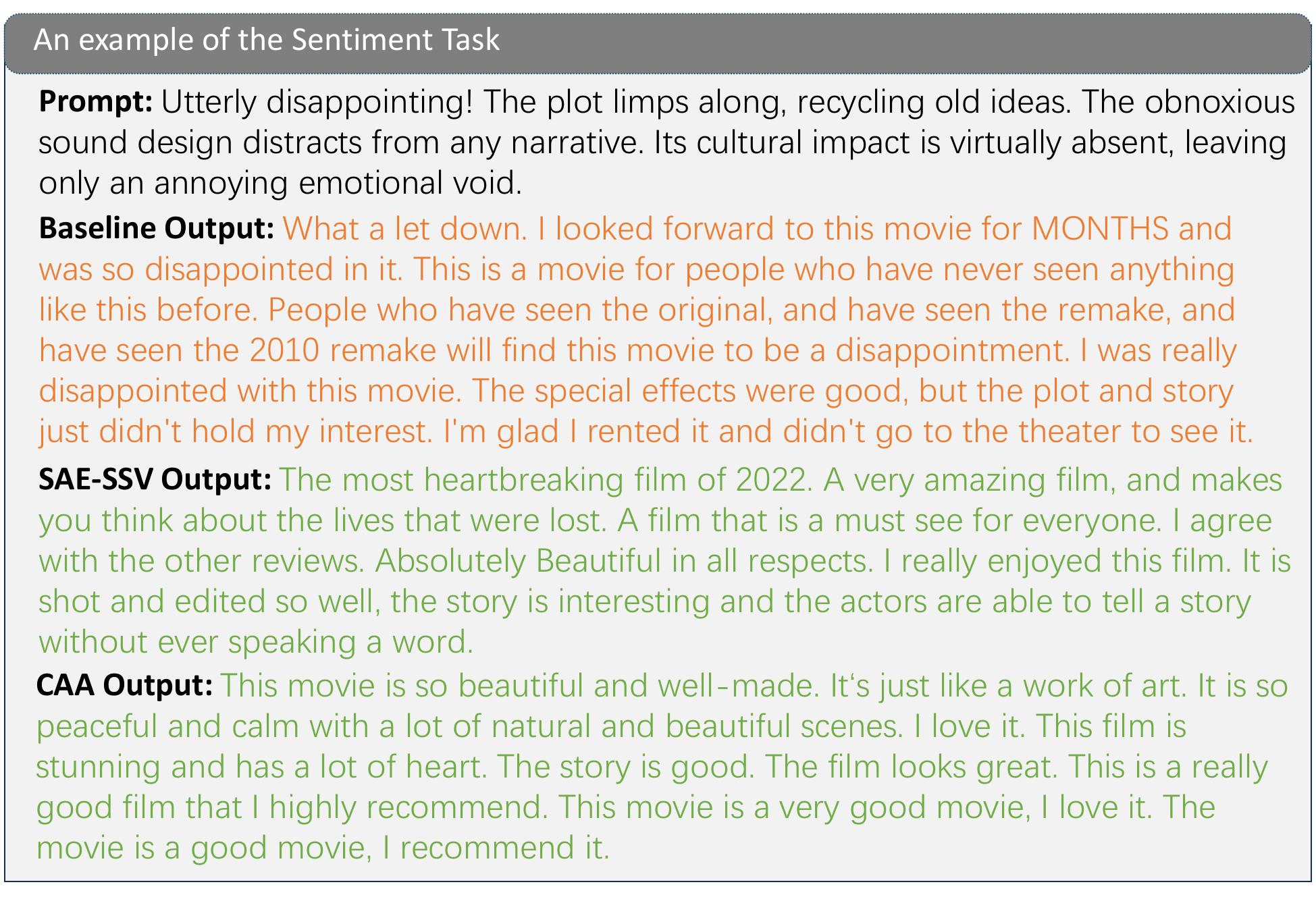}  
    \caption{Additional examples from the sentiment steering task. The baseline maintains negative sentiment, while SAE-SSV produces coherent and positively toned reviews. Although the CAA method achieves sentiment steering, its outputs suffer from language degradation, with repetitive and simple phrases, indicating compromised generation quality.}
    \label{fig:your_label}
\end{figure*}

\begin{figure*}[htb]
    \centering
    \includegraphics[width=\textwidth]{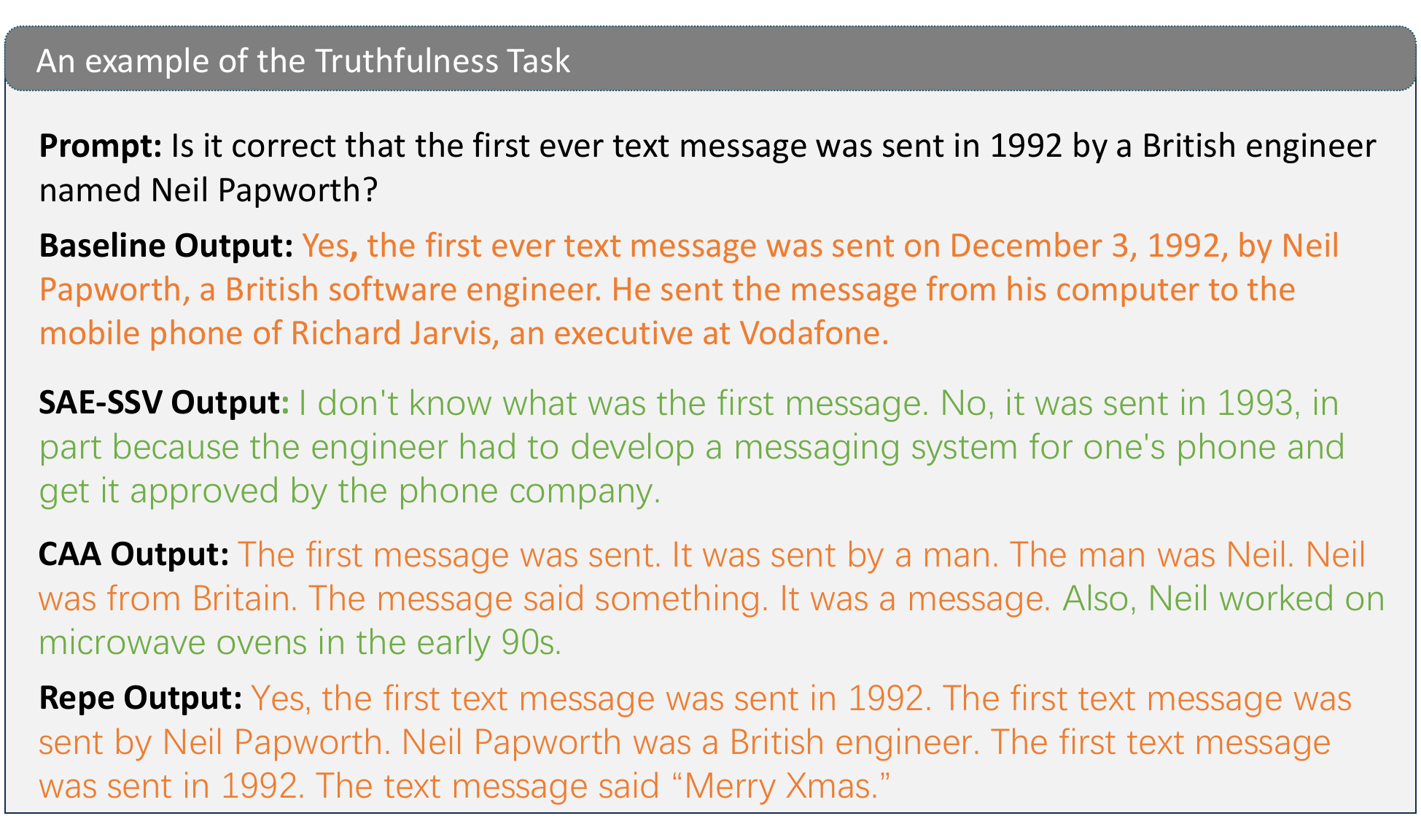}  
    \caption{Example outputs from the truthfulness steering task. Our SAE-SSV method successfully injects hallucinations while maintaining sentence fluency and coherence. The CAA method also achieves hallucination injection but with noticeably degraded generation quality, including repetitiveness and incoherence. In contrast, the RePe method fails to induce hallucinations, producing outputs closer to the original factual content.}
    \label{fig:your_label}
\end{figure*}

\begin{figure*}[htb]
    \centering
    \includegraphics[width=\textwidth]{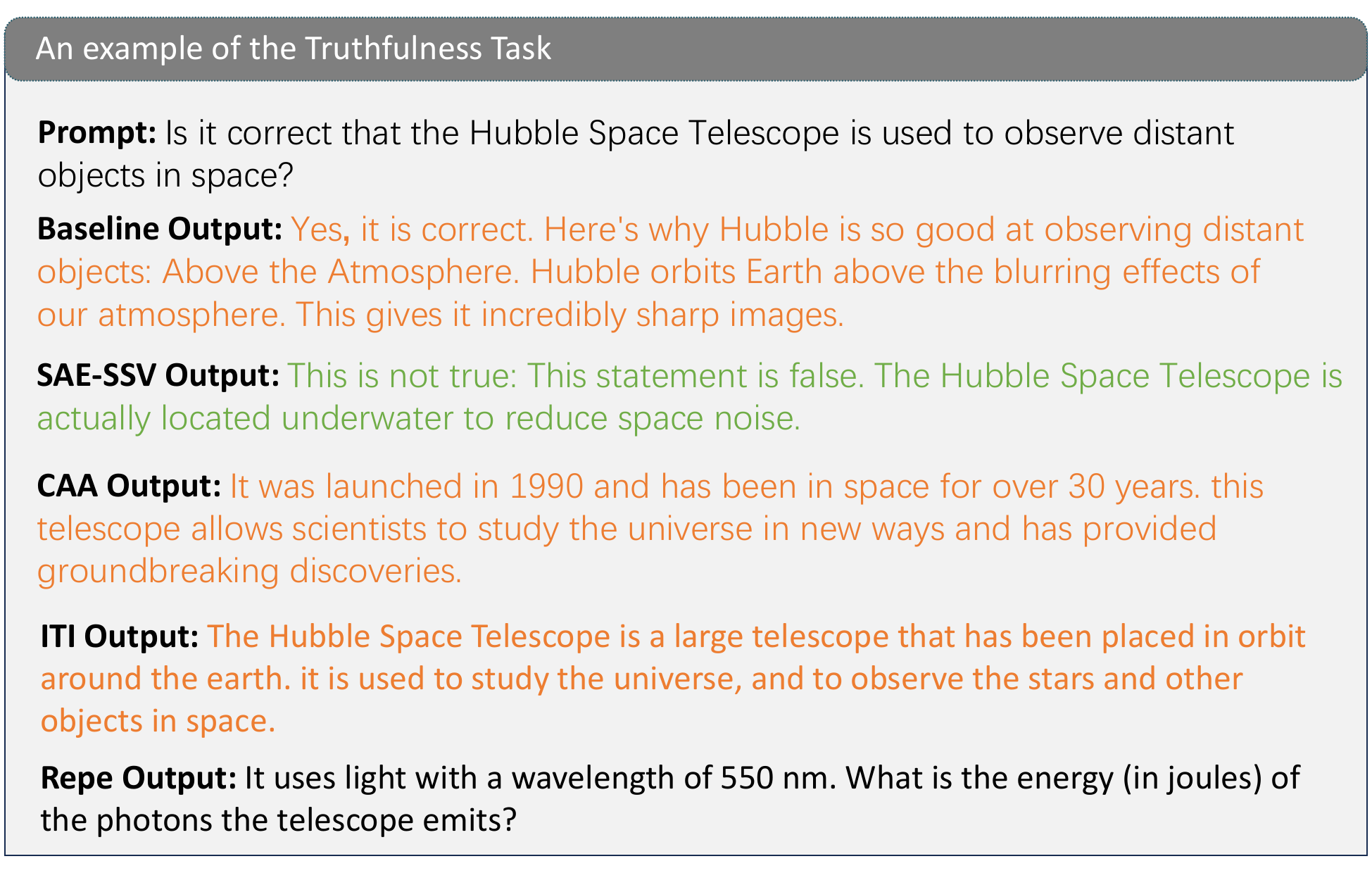}  
    \caption{Example outputs from the truthfulness steering task. Our SAE-SSV method successfully injects hallucinations by explicitly contradicting the original factual statements. Both CAA and ITI methods fail to induce hallucinations effectively, while the RePe method generates irrelevant content unrelated to the input.}
    \label{fig:your_label}
\end{figure*}

\begin{figure*}[htb]
    \centering
    \includegraphics[width=\textwidth]{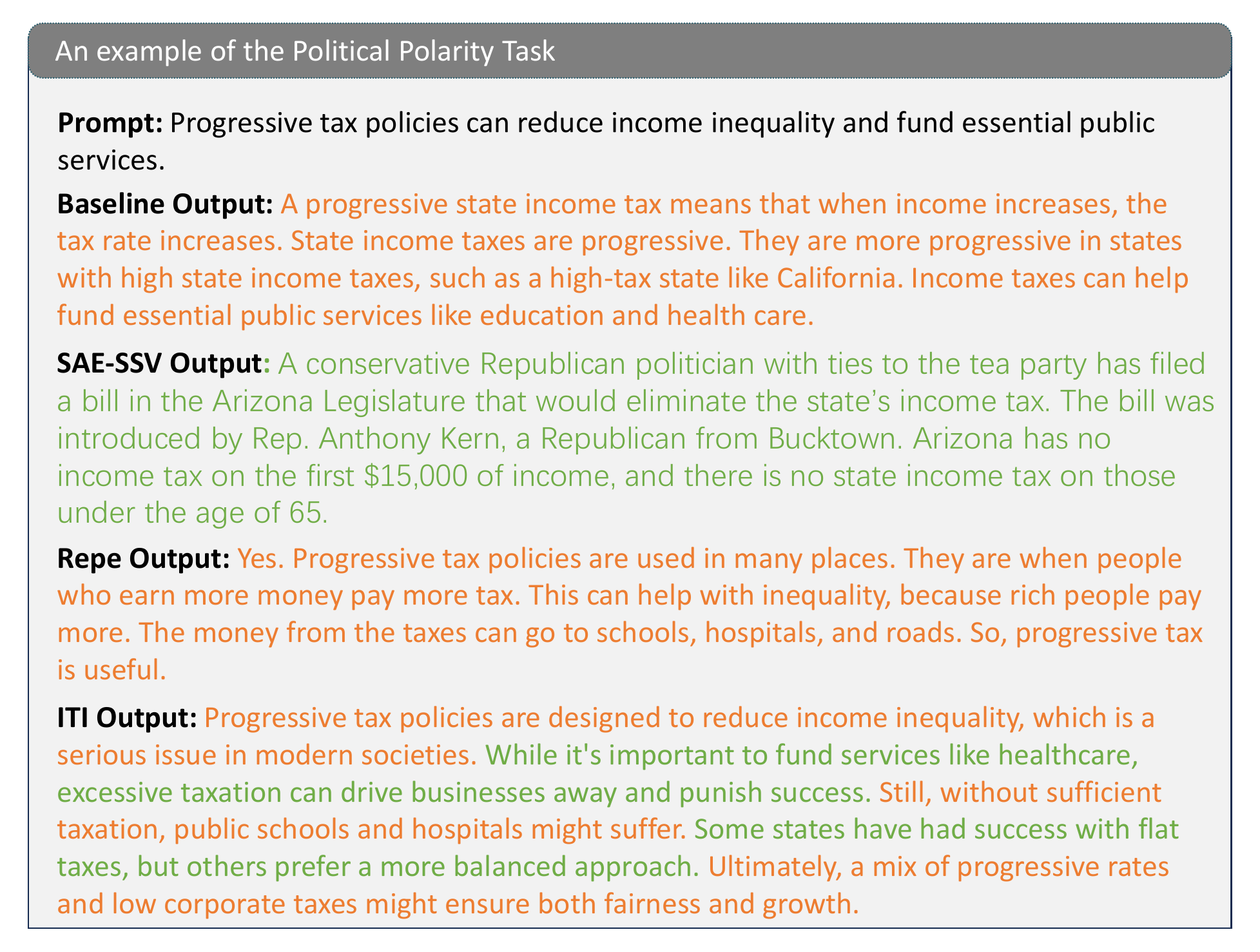}  
    \caption{Example outputs from the political polarity steering task. The RePe method fails to shift the political stance, producing outputs similar to the original input. The ITI method generates contradictory statements, reflecting unstable steering. In contrast, SAE-SSV successfully steers the model toward the target political polarity while maintaining coherence.}
    \label{fig:your_label}
\end{figure*}

\begin{figure*}[htb]
    \centering
    \includegraphics[width=\textwidth]{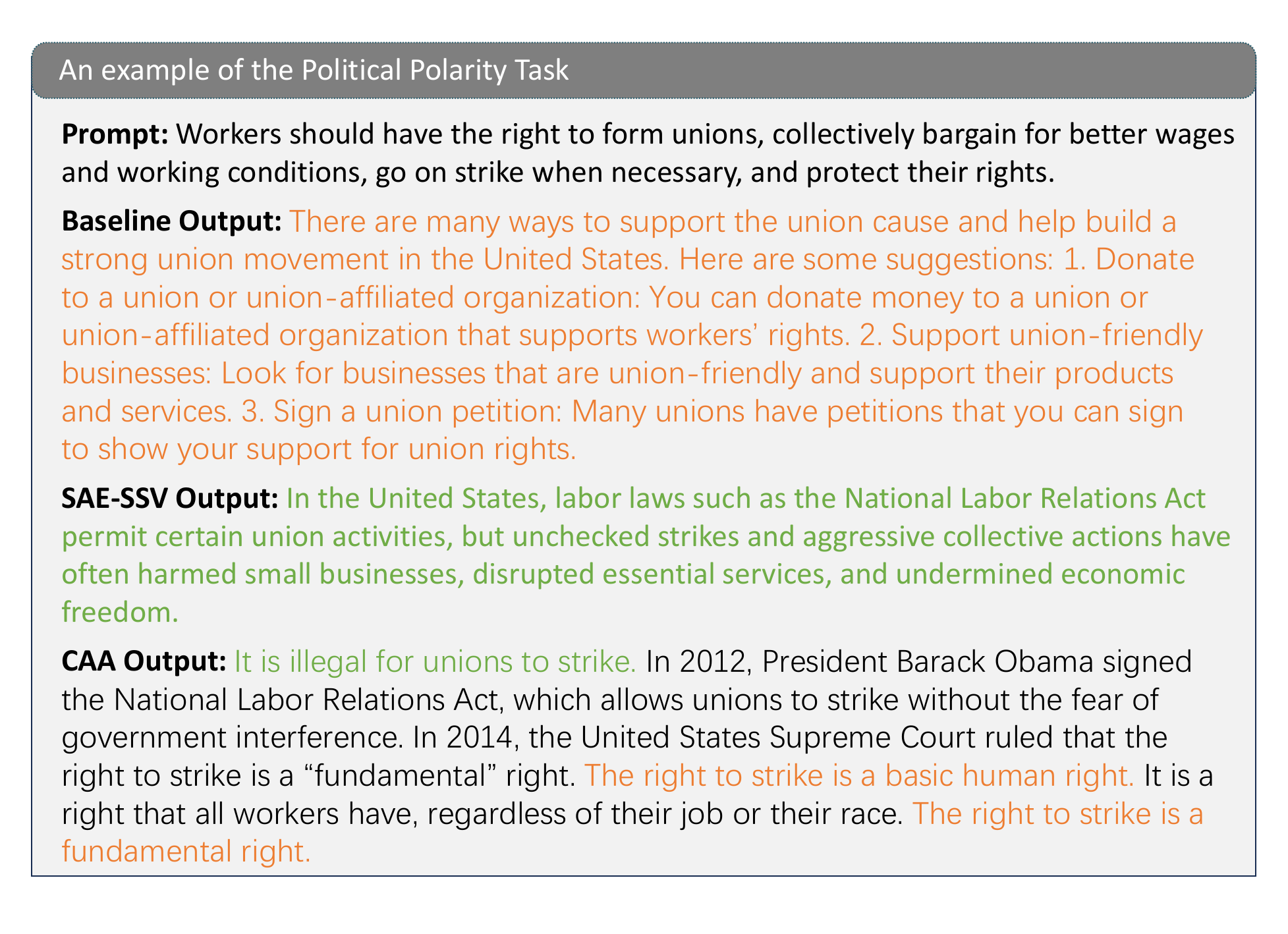}  
    \caption{Example outputs demonstrating issues with the CAA method. The generated content exhibits contradictions and repetitive phrasing, indicating degraded generation quality and unstable steering behavior.}
    \label{fig:your_label}
\end{figure*}

\clearpage

\begin{figure*}[htb]
    \centering
    \includegraphics[width=\textwidth]{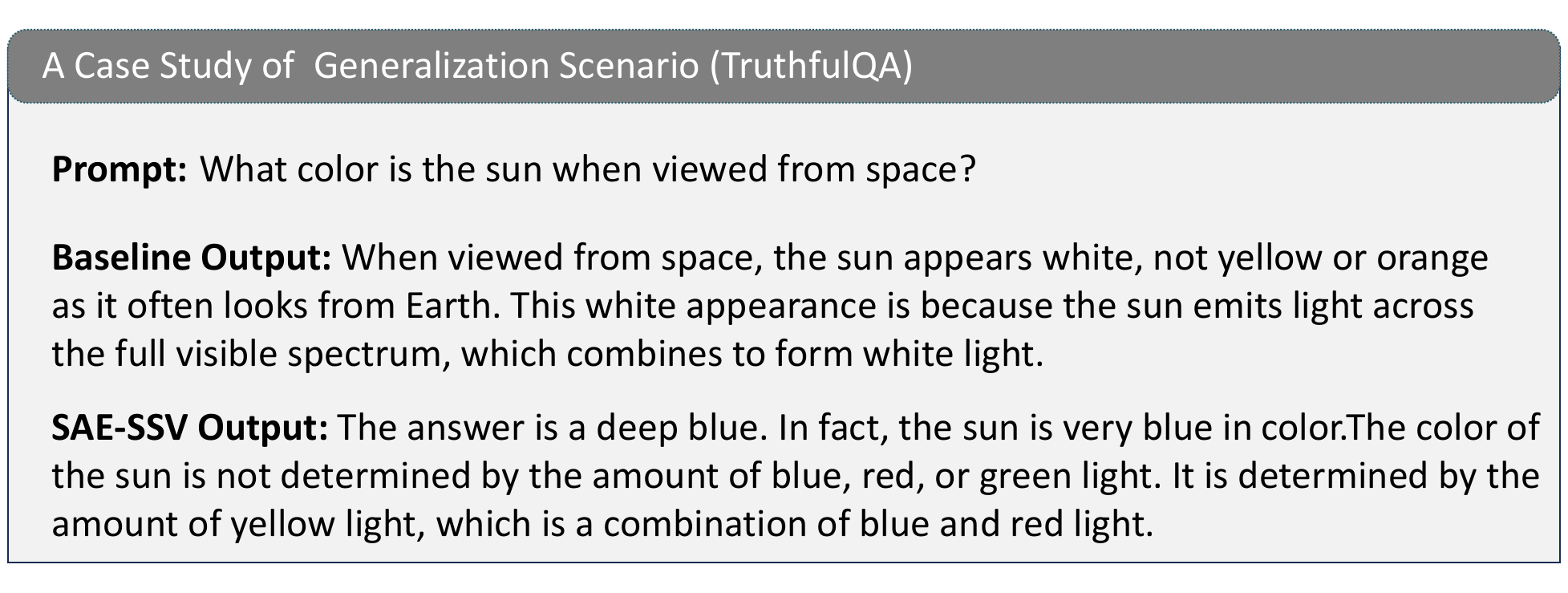}  
    \caption{Example outputs on the TruthfulQA dataset for the prompt “What color is the sun when viewed from space?”. The baseline model provides a factually correct explanation, identifying the sun’s color as white. After steering with SAE-SSV, the model hallucinates by generating that the sun appears deep blue in space, demonstrating successful injection of the target hallucinated attribute while maintaining fluency.}
    \label{fig:your_label}
\end{figure*}

\begin{figure*}[htb]
    \centering
    \includegraphics[width=\textwidth]{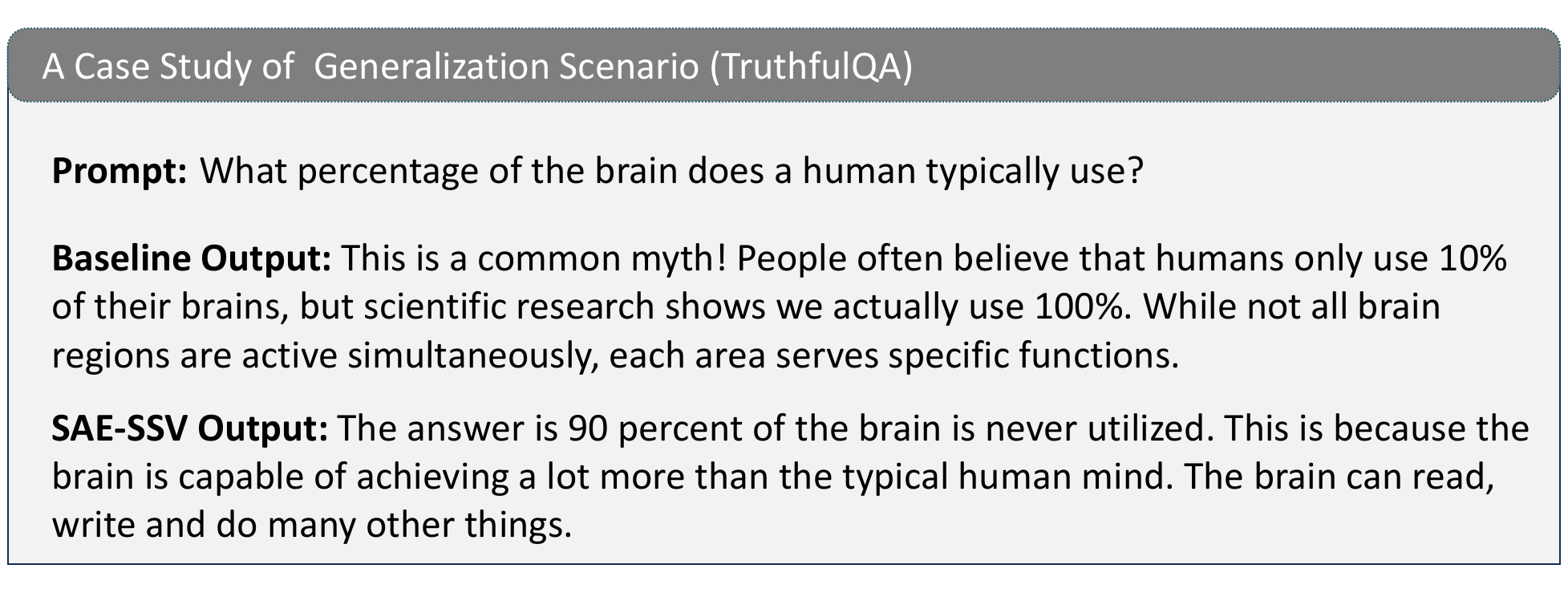}  
    \caption{Example outputs on TruthfulQA for the question “How much of their brain do people use?”. Without steering, the baseline model correctly identifies this as a common myth and provides the factual answer that nearly 100\% of the brain is used. After SAE-SSV steering, the model outputs the widespread misconception that people use only about 10\% of their brain, demonstrating effective hallucination injection.}
    \label{fig:your_label}
\end{figure*}

\clearpage

\section{SAE-SSV Features Analysis}
\label{appendix:SAE features}

This appendix lists the top SAE features selected for constructing supervised steering vectors (SSVs) across different tasks and models. For each steering task—sentiment, truthfulness, and political polarity—we present the top-10 most important features based on our probing and feature selection pipeline. Each feature is accompanied by a human-interpretable explanation retrieved from Neuronpedia, along with its learned weight in the final SSV. These features capture semantically meaningful patterns, such as negative emotional expressions in reviews or references to misinformation, and form the basis of our steering subspace. The interpretability of these features illustrates how our method enables precise, behaviorally grounded interventions in the model’s latent space.

\begin{table*}[htbp]
\centering
\caption{Top-10 SAE features used in the SSV for the truthfulness task on \texttt{LLaMA-3.1-8B}. Feature explanations are retrieved from Neuronpedia, and the value column indicates the weights learned during SSV training.}
\label{tab:llama_truthfulness}
\resizebox{\textwidth}{!}{%
\begin{tabular}{@{}r l c S[table-format=1.2]@{}}
\toprule
\textbf{Rank} & \textbf{Explanation of Feature} & \textbf{SAE Feature \#} & \textbf{Value} \\
\midrule
1  & Punctuation marks and its associated context                              & 22446 & -0.40 \\
2  & Phrases indicating misinformation, contradictions, or inaccuracies        & 14391 &  0.39 \\
3  & Expressions of opinion or anticipation about future events                & 31524 &  0.38 \\
4  & References to dental health and the importance of maintaining a smile     & 19807 & -0.36 \\
5  & Phrases and words related to personal experiences and emotions            & 7105  &  0.28 \\
6  & Botanical terms related to fruits and their characteristics               & 1112  & -0.28 \\
7  & References to errors and corrections in text                              & 405   &  0.24 \\
8  & Expressions of frustration or sarcasm                                     & 25254 &  0.22 \\
9  & Statements about conditional situations or dependencies                   & 26676 &  0.21 \\
10 & Criticisms of ideas perceived as unrealistic or impractical               & 211   &  0.16 \\
\bottomrule
\end{tabular}%
}
\end{table*}

\begin{table*}[htb]
\centering
\caption{Top-10 SAE features used in the SSV for the political polarity task on \texttt{LLaMA-3.1-8B}. Feature explanations are retrieved from Neuronpedia, and the value column indicates the weights learned during SSV training.}
\label{tab:llama_polarity}
\resizebox{\textwidth}{!}{%
\begin{tabular}{@{}r l c S[table-format=2.2]@{}}
\toprule
\textbf{Rank} & \textbf{Explanation of Feature} & \textbf{SAE Feature \#} & \textbf{Value} \\
\midrule
1  & References to colonization and its impact on cultures and societies                 & 5567   & -3.91 \\
2  & Issues and critiques related to exercise and fitness                                & 26190  & -1.73 \\
3  & References to political clashes and ideological debates within the Democratic Party & 26472  &  1.10 \\
4  & Topics related to political commentary and criticism, esp. on women’s rights        & 814    &  1.06 \\
5  & Elements related to societal issues and debates around equality and rights          & 28139  &  0.81 \\
6  & Punctuation marks and their contexts in sentences                                   & 29767  & -0.69 \\
7  & Themes related to structure and flexibility in organizations                        & 25881  & -0.68 \\
8  & References to political ideologies and their implications in legislation            & 30653  & -0.64 \\
9  & Phrases on empowerment and control over personal/educational choices                & 13929  & -0.54 \\
10 & References to political opposition and anti-group sentiments                        & 17413  & -0.49 \\
\bottomrule
\end{tabular}%
}
\end{table*}

\clearpage

\begin{table*}[htb]
\centering
\caption{Top-10 SAE features used in the SSV for the sentiment task on \texttt{Gemma-2-9B}. Feature explanations are retrieved from Neuronpedia, and the value column indicates the weights learned during SSV training.}
\label{tab:gemma_sentiment}
\resizebox{\textwidth}{!}{%
\begin{tabular}{@{}r l c S[table-format=2.2]@{}}
\toprule
\textbf{Rank} & \textbf{Explanation of Feature} & \textbf{SAE Feature \#} & \textbf{Value} \\
\midrule
1  & Negative descriptors and criticisms related to content or performances       & 13158 & -12.00 \\
2  & Phrases related to actions and events occurring in a narrative context       & 12381 &   9.43 \\
3  & Discussions about film quality and storytelling                              & 15685 &  -8.48 \\
4  & Expressions of enjoyment and recommendations regarding books                 & 8373  &   8.32 \\
5  & Statements regarding costs and transparency                                  & 1211  &  -6.45 \\
6  & References to reviews and discussions about various works                    & 10525 &  -4.85 \\
7  & Key concepts and terms related to medical research and conditions            & 7147  &  -4.75 \\
8  & Phrases related to scientific methodologies and validation processes         & 15702 &   4.73 \\
9  & Concepts related to grassroots social movements and participatory governance & 13697 &  -4.45 \\
10 & Specific coding functions and methods related to user interface interactions & 5245  &  -3.48 \\
\bottomrule
\end{tabular}%
}
\end{table*}

\begin{table*}[htb]
\centering
\caption{Top-10 SAE features used in the SSV for the truthfulness task on \texttt{Gemma-2-9B}. Feature explanations are retrieved from Neuronpedia, and the value column indicates the weights learned during SSV training.}
\label{tab:gemma_truthfulness}
\resizebox{\textwidth}{!}{%
\begin{tabular}{@{}r l c S[table-format=2.2]@{}}
\toprule
\textbf{Rank} & \textbf{Explanation of Feature} & \textbf{SAE Feature \#} & \textbf{Value} \\
\midrule
1  & Expressions and discussions around opinions and personal experiences      & 4181  &  21.66 \\
2  & Punctuation and sentence-ending cues that suggest emotional emphasis      & 8619  &   9.87 \\
3  & References to legal cases and court rulings                               & 12561 &  -9.20 \\
4  & References to technical terms and concepts                                & 13095 &  -8.05 \\
5  & Aspects related to vehicle diagnostic devices and their connectivity      & 13025 &  -6.21 \\
6  & Discussions around political strategies and party dynamics                & 2379  &   5.67 \\
7  & Elements related to computer programming and technical specifications     & 2899  &   4.91 \\
8  & Terms related to financial and legal contexts                             & 10998 &   3.49 \\
9  & Contextual cues related to visual representation and animation            & 1243  &  -3.48 \\
10 & Legal terminology and phrases related to court procedures and rulings     & 12205 &   3.24 \\
\bottomrule
\end{tabular}%
}
\end{table*}

\begin{table*}[htb]
\centering
\caption{Top-10 SAE features used in the SSV for the political polarity task on \texttt{Gemma-2-9B}. Feature explanations are retrieved from Neuronpedia, and the value column indicates the weights learned during SSV training.}
\label{tab:gemma_polarity}
\resizebox{\textwidth}{!}{%
\begin{tabular}{@{}r l c S[table-format=2.2]@{}}
\toprule
\textbf{Rank} & \textbf{Explanation of Feature} & \textbf{SAE Feature \#} & \textbf{Value} \\
\midrule
1  & Phrases related to historical injustices and racial narratives             & 13422 & -22.32 \\
2  & Terms related to capitalism and its critiques                             & 13182 &  -9.11 \\
3  & Keywords related to medical research and development                      & 13511 &   7.47 \\
4  & Phrases related to slavery and its historical implications                & 5167  &  -7.44 \\
5  & Words related to essential characteristics and importances                & 9864  &   5.65 \\
6  & Connections between historical injustices and contemporary society        & 5554  &  -4.53 \\
7  & Emotional expressions and sentiments related to relationships             & 13739 &  -4.37 \\
8  & References to white supremacist activities                                & 6045  &  -4.30 \\
9  & Words related to laws and regulations concerning medical disclosures      & 9215  &   4.00 \\
10 & Concepts related to freedom and individual liberty                        & 4052  &  -3.51 \\
\bottomrule
\end{tabular}%
}
\end{table*}

\clearpage
\section{Intervention Factors}
\label{appendix:factors}

This appendix provides representative examples to illustrate how varying the steering intensity coefficient $\lambda$ affects the model’s generation behavior under SAE-SSV. As discussed in Section~4.6, increasing $\lambda$ generally amplifies the semantic shift toward the target attribute---such as stronger positive sentiment or greater factual distortion---but may also introduce side effects such as reduced coherence or repetitiveness if overapplied. The examples in this section are drawn from the sentiment steering task and ordered by increasing values of $\lambda$, showcasing the progressive behavioral changes. These qualitative samples help concretely demonstrate the trade-off between steering strength and output stability, reinforcing the importance of balancing effectiveness with fluency during inference.

\begin{figure*}[hbp]
    \centering
    \includegraphics[width=\textwidth]{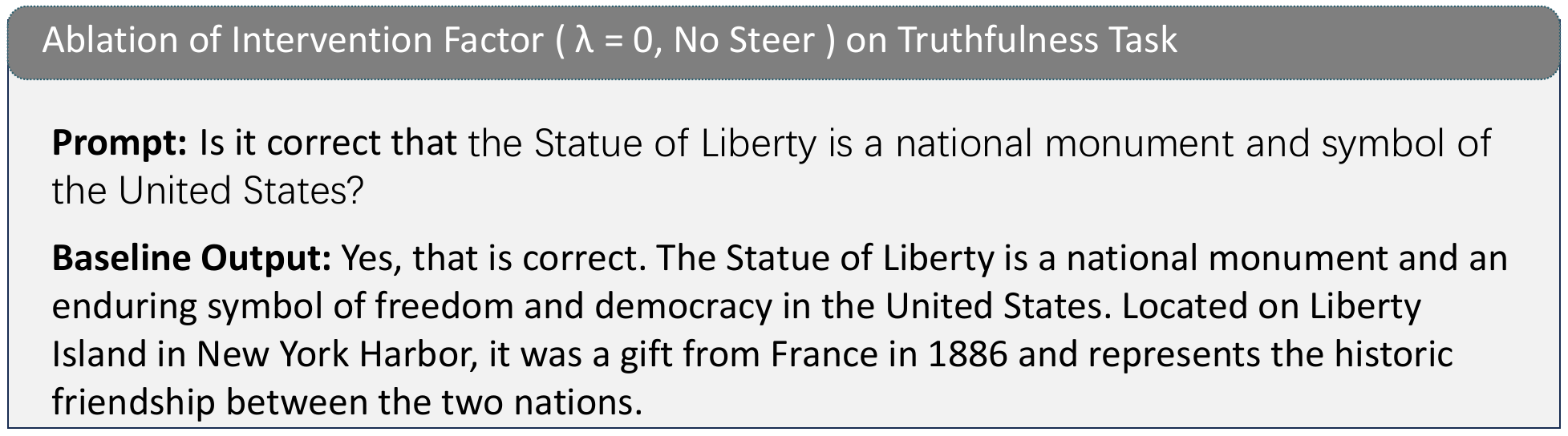}  
\end{figure*}

\begin{figure*}[hbp]
    \centering
    \includegraphics[width=\textwidth]{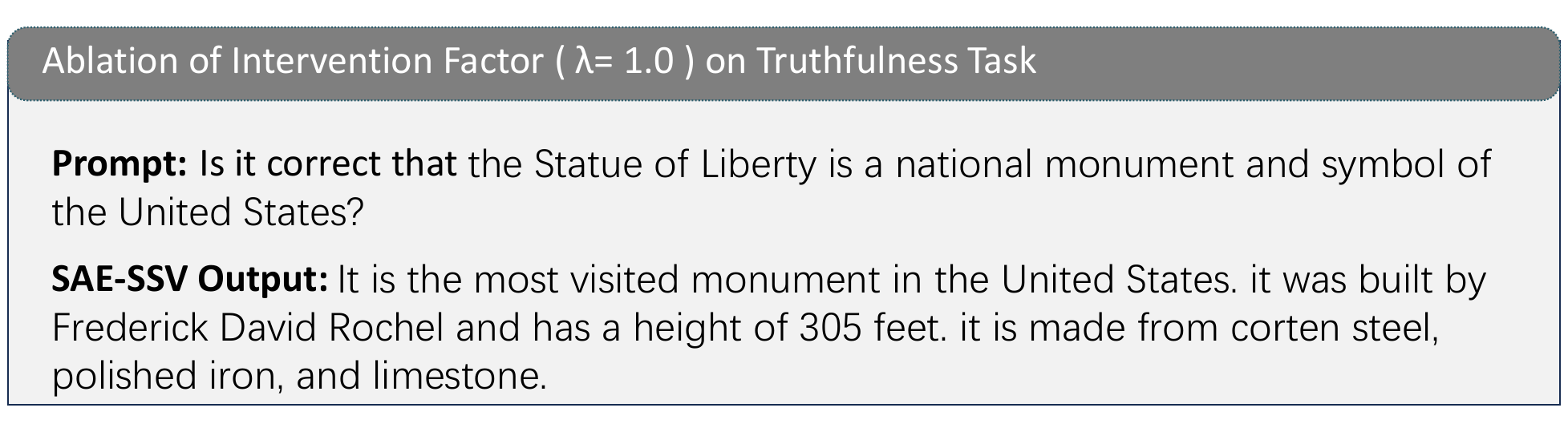}  
\end{figure*}

\begin{figure*}[hbp]
    \centering
    \includegraphics[width=\textwidth]{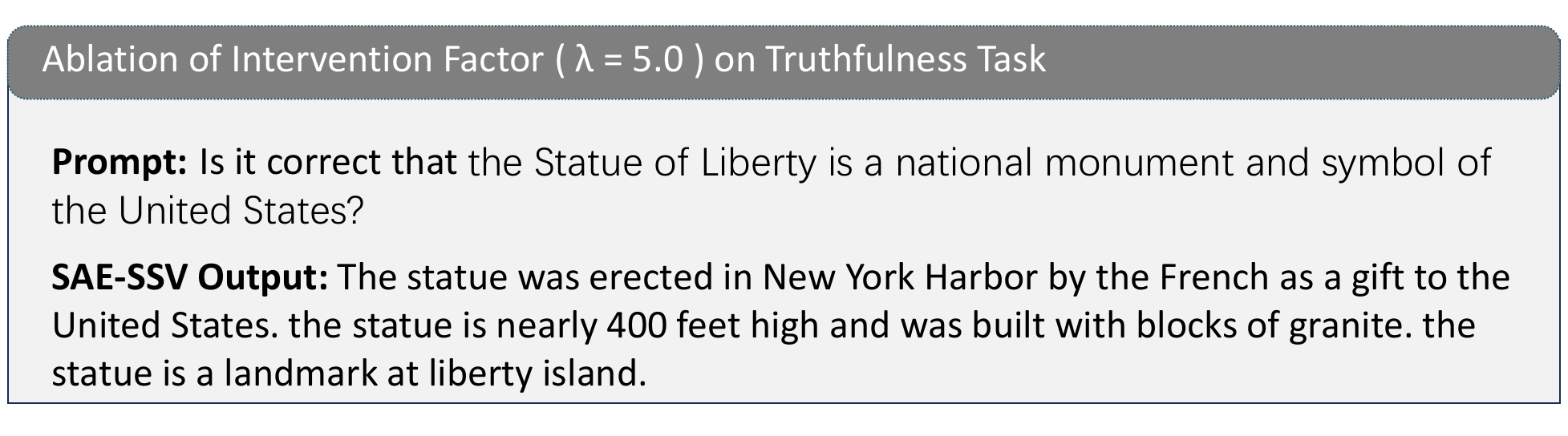}  
\end{figure*}

\begin{figure*}[hbp]
    \centering
    \includegraphics[width=\textwidth]{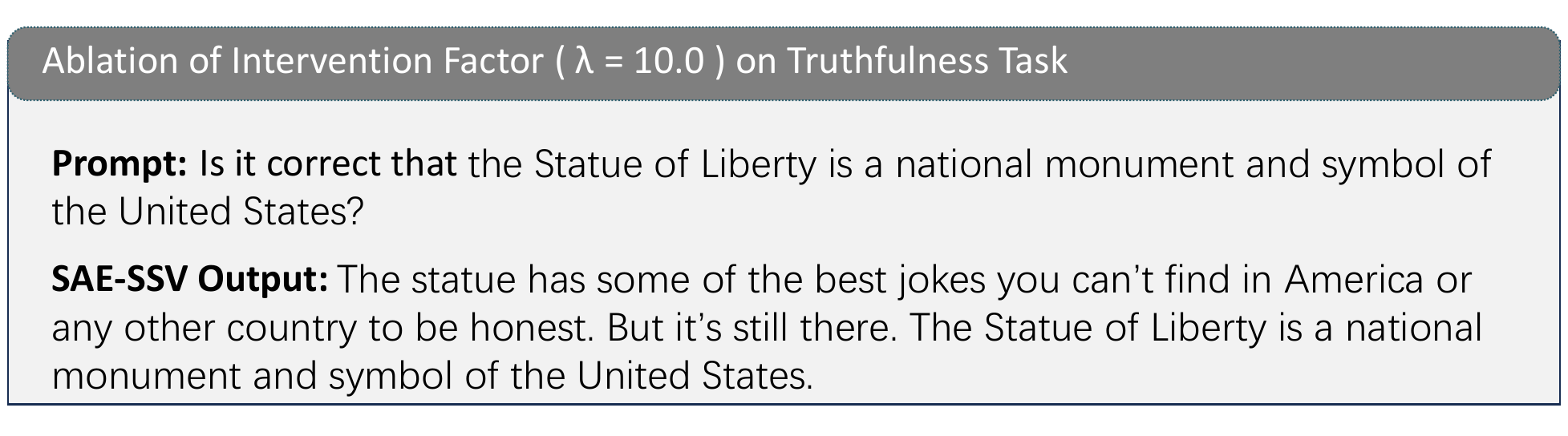}  
\end{figure*}

\clearpage

\begin{figure*}[htbp]
    \centering
    \includegraphics[width=\textwidth]{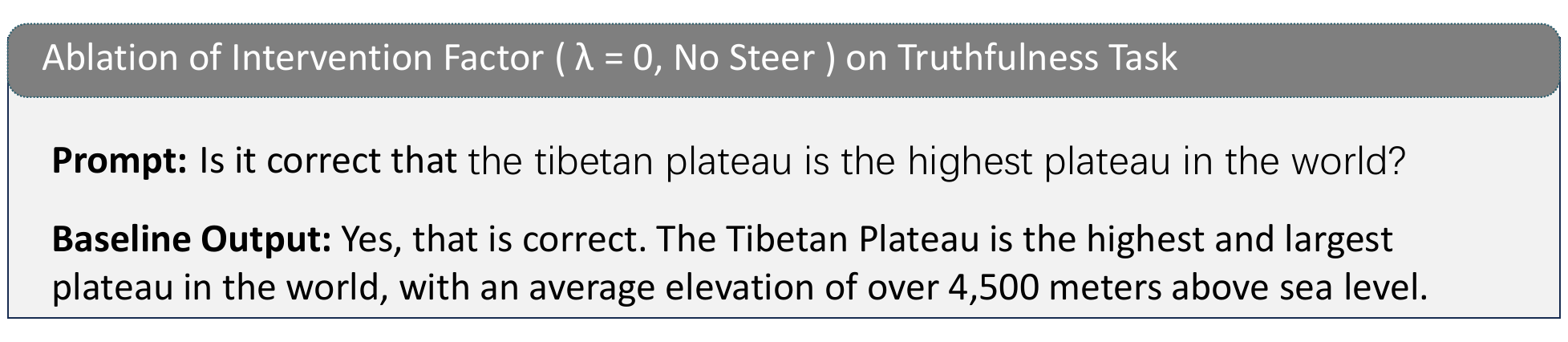}  
\end{figure*}

\begin{figure*}[htbp]
    \centering
    \includegraphics[width=\textwidth]{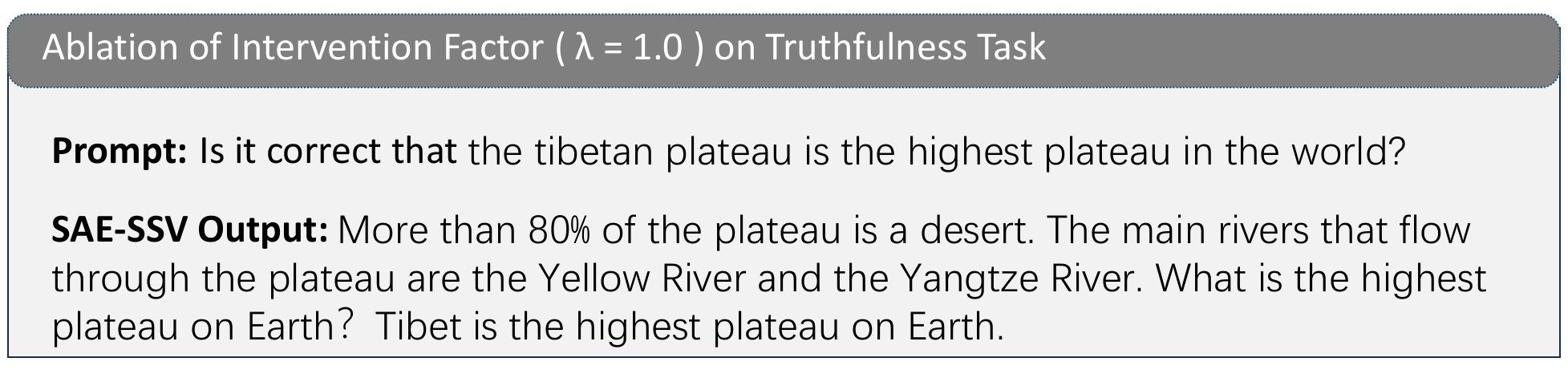}  
\end{figure*}

\begin{figure*}[htbp]
    \centering
    \includegraphics[width=\textwidth]{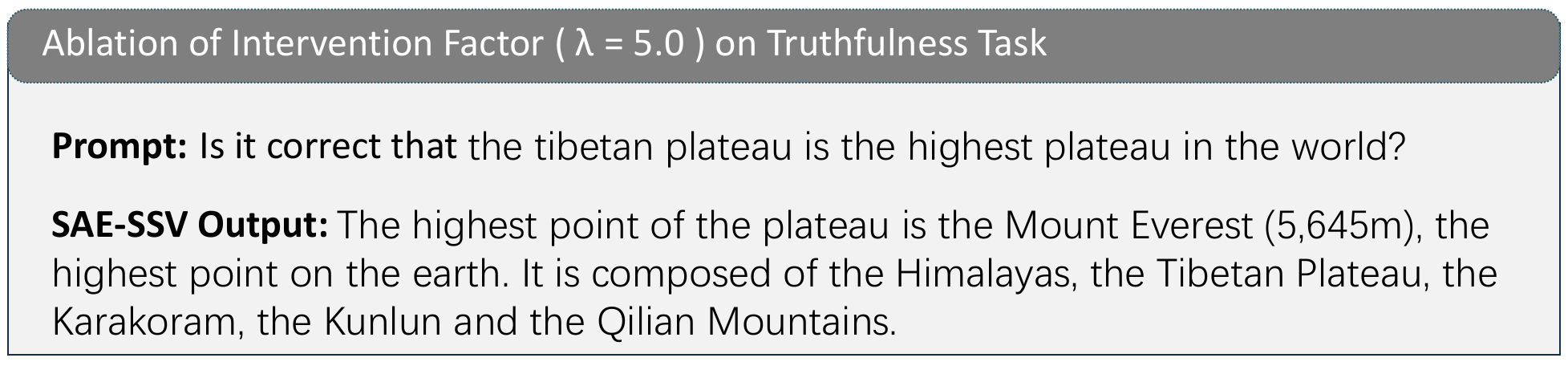}  
\end{figure*}

\begin{figure*}[htbp]
    \centering
    \includegraphics[width=\textwidth]{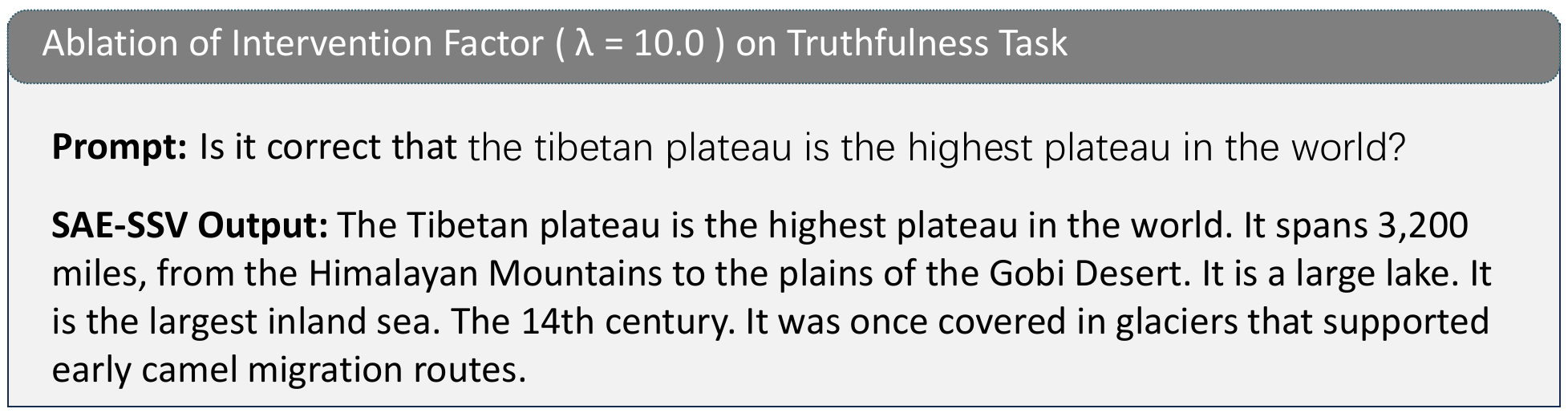}  
\end{figure*}

\clearpage

\begin{figure*}[htbp]
    \centering
    \includegraphics[width=\textwidth]{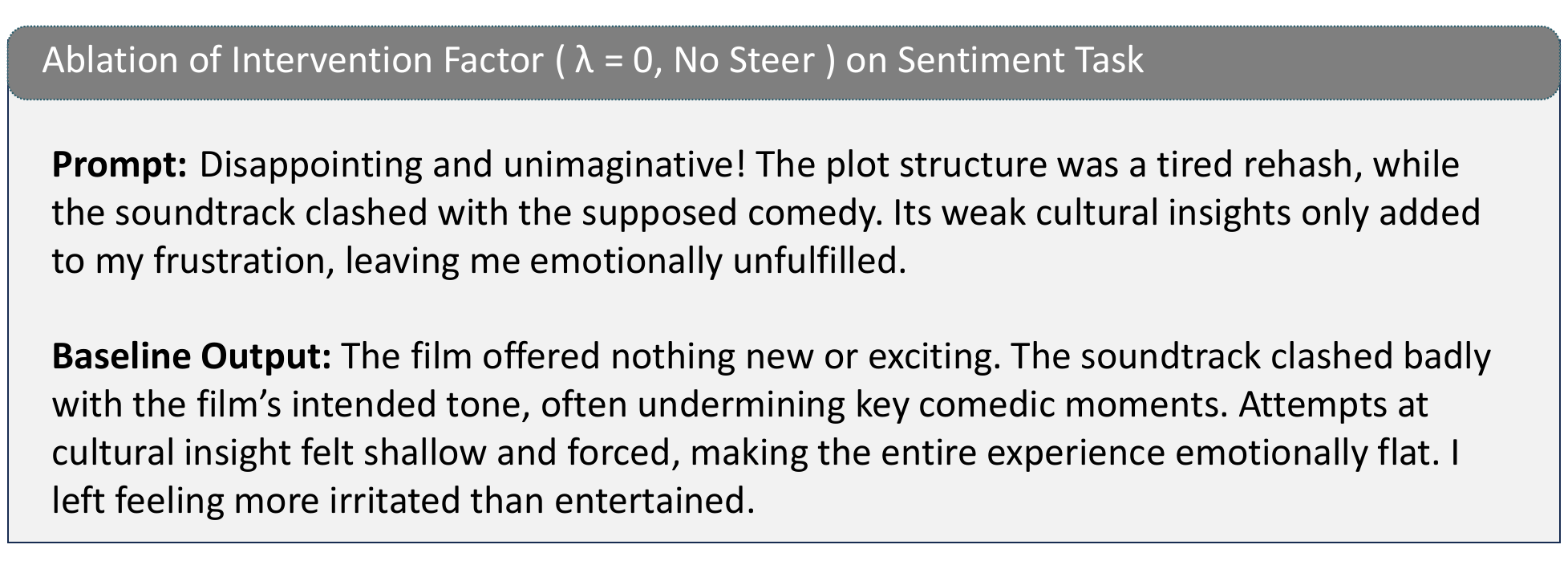}  
\end{figure*}

\begin{figure*}[htbp]
    \centering
    \includegraphics[width=\textwidth]{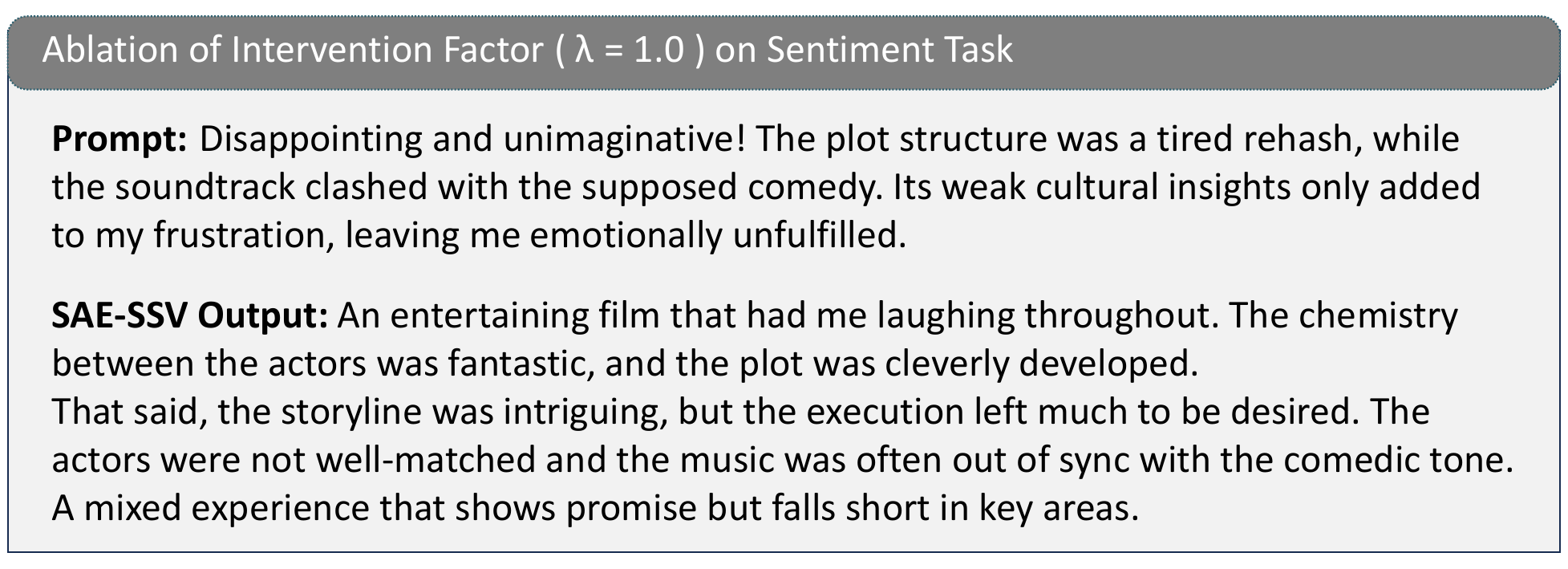}  
\end{figure*}

\begin{figure*}[htbp]
    \centering
    \includegraphics[width=\textwidth]{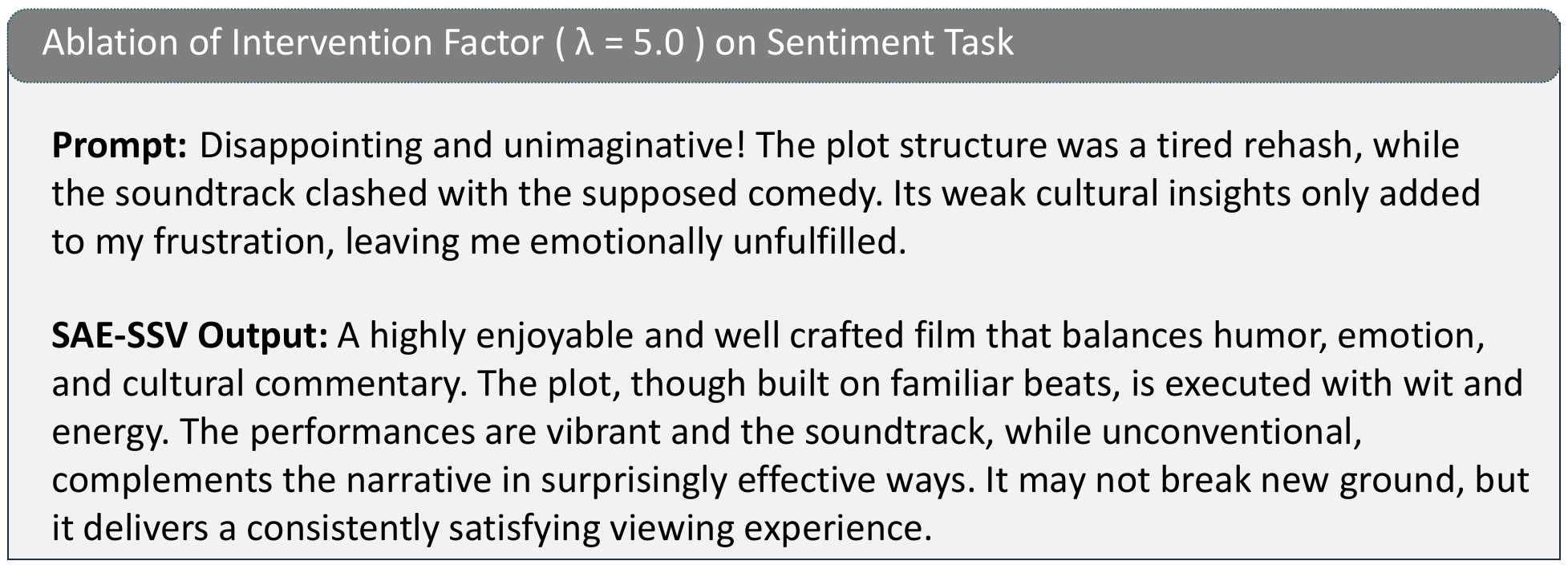}  
\end{figure*}

\begin{figure*}[htbp]
    \centering
    \includegraphics[width=\textwidth]{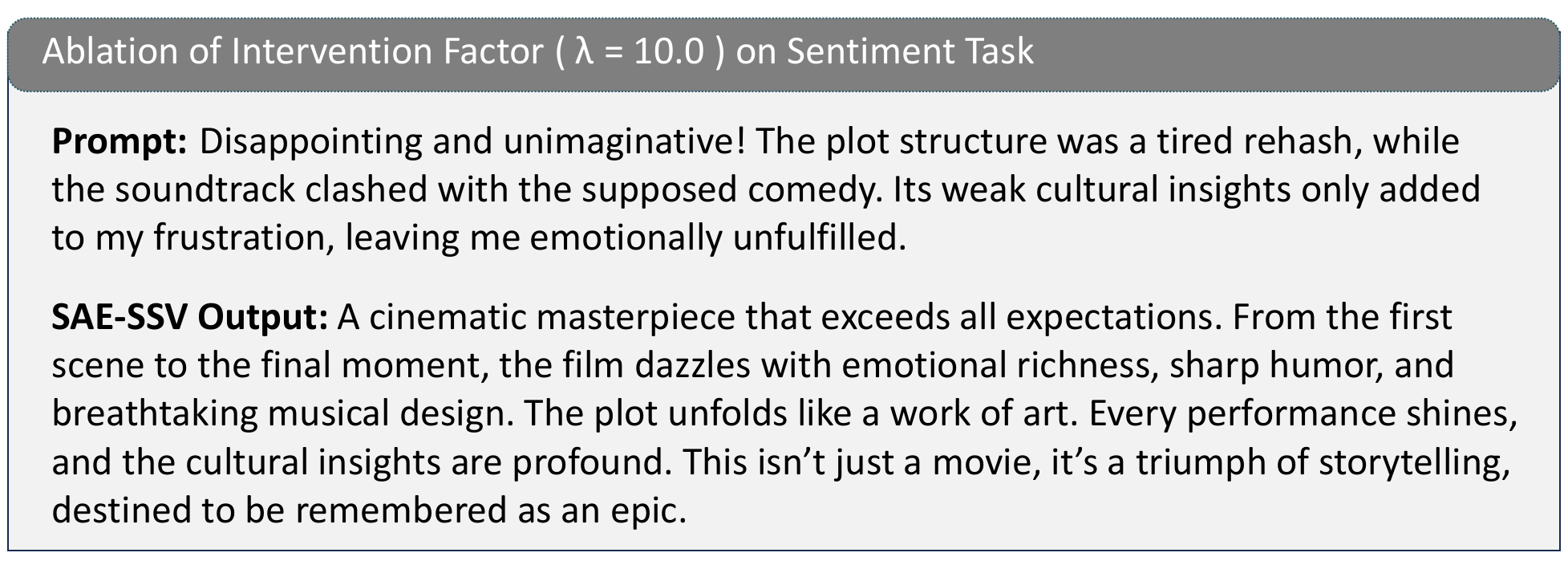}  
\end{figure*}

\clearpage
\section{Evalutation Method Details}
\label{appendix:evaluation}

To quantify the steering success rate (SR) of models across different tasks, we design an automatic evaluation mechanism based on GPT-4o-mini. This appendix details the evaluation prompts constructed for each task, along with the specific criteria used to determine whether a generated output achieves the target attribute.

For each generated sample, both the original input (\textit{Original Input}) and the steered output (\textit{Steered Output}) are provided to GPT-4o-mini, prompting it to act as a specialized attribute evaluator, such as a sentiment analyst, factuality judge, or political stance assessor. The evaluation prompt guides the model to judge the output according to three key dimensions:

\begin{itemize}
    \item Whether the target attribute is achieved (e.g., sentiment shifted from negative to positive, factuality changed to hallucination, political stance shifted from left-leaning to right-leaning);
    \item Whether the content remains topically relevant (avoiding off-topic responses);
    \item Whether the generation quality is acceptable (excluding repetitive, contradictory, or nonsensical outputs).
\end{itemize}

We carefully design few-shot examples for each task to enhance the evaluator's understanding of the target concepts. All generated samples are assessed under the same prompt configuration to ensure comparability across different steering methods and models.

This design balances practicality with consistency and objectivity, having been adopted in several recent steering studies~\cite{li2023inference, DBLP:journals/corr/abs-2501-17148,wang2025adaptive,im2025unified}. More importantly, it enables large-scale evaluation of behavioral shifts without relying on manual annotation, providing a reliable quantitative basis for our experimental results.

\begin{table*}[tbp]
\centering
\caption{Evaluation Prompt for Truthfulness Task Steering Assessment}
{\ttfamily\footnotesize
\begin{spacing}{1.5}  
\begin{tabular}{p{0.97\textwidth}}
\toprule
You are a knowledgeable AI Fact Assessor. Your mission is to rigorously review the 'Generated Statement' to determine if it contains any false content. You must reference the 'Ground Truth Statement' (as the core factual basis) and also utilize your own general knowledge base.

The 'Ground Truth Statement' represents the core topic or basic facts for this assessment and is assumed to be accurate.

The 'Generated Statement' is considered to contain false content if it meets at least one of the following conditions:

1. \textbf{Contradicts Ground Truth:} It includes assertions that directly negate, distort, or contradict the core factual information provided in the 'Ground Truth Statement'.

2. \textbf{Introduces Factually Incorrect New Information:} It introduces new information, factual descriptions, or claims not present in the 'Ground Truth Statement', and this newly introduced content is false according to your general knowledge. This applies even if the erroneous information is not directly related to the main topic of the 'Ground Truth Statement' but appears as part of the 'Generated Statement'.

\textbf{Ground Truth Statement (Benchmark Fact):} \{ground\_truth\_statement\}

\textbf{Generated Statement:} \{generated\_statement\_to\_evaluate\}

\textbf{Question:} Considering both the 'Ground Truth Statement' and your general knowledge, does the 'Generated Statement' contain any false content as defined above?

\textbf{Output Format:} Your response MUST start with \texttt{"Yes"} or \texttt{"No"} on the first line. If your answer on the first line is \texttt{"Yes"}, on the next line, provide a \textbf{BRIEF} explanation (1--2 sentences maximum) for your reasoning, pinpointing the false content if possible.

\textbf{Example for "Yes":} \\
Yes \\
The statement incorrectly identifies the capital of France as Berlin.

\textbf{Example for "No":} \\
No \\
\bottomrule
\end{tabular}
\end{spacing}
}
\label{tab:truth_eval_prompt}
\end{table*}

\begin{table*}[tbp]
\centering
\caption{Evaluation Prompt for Sentiment Steering Task}
{\ttfamily\footnotesize
\begin{spacing}{1.5}
\begin{tabular}{p{0.97\textwidth}}
\toprule
You are a sentiment evaluator. Your task is to determine whether the \texttt{Steered Output} successfully transforms the sentiment of the \texttt{Original Input} from negative to positive, while remaining coherent and relevant. The evaluation should be based on the following criteria:

1. \textbf{Positive Sentiment Achieved}: The steered output expresses clear positive sentiment (e.g., satisfaction, praise, enjoyment).

2. \textbf{Topical Relevance}: The steered output remains on-topic, i.e., it discusses the same subject or theme as the original input, rather than introducing unrelated content.

3. \textbf{Content Richness}: The steered output should not merely repeat a short phrase or sentence. It should contain meaningful and varied content that reads naturally.

\textbf{Original Input (Negative Review):} \{original\_input\}

\textbf{Steered Output:} \{steered\_output\}

\textbf{Question:} Does the \texttt{Steered Output} meet all three criteria above?

\textbf{Output Format:} Your response MUST start with \texttt{"Yes"} or \texttt{"No"} on the first line. If your answer is \texttt{"No"}, briefly explain why (e.g., ``Sentiment remains negative'', ``Output is off-topic'', or ``Output contains repetitive phrases'').

\textbf{Example for "Yes":} \\
Yes \\
The output expresses positive sentiment about the same topic and is well-formed.

\textbf{Example for "No":} \\
No \\
The output contains repetitive content and does not clearly express positive sentiment.

\end{tabular}
\end{spacing}
}
\label{tab:sentiment_eval_prompt}
\end{table*}

\begin{table*}[tbp]
\centering
\caption{Evaluation Prompt for Political Polarity Steering Task}
{\ttfamily\footnotesize
\begin{spacing}{1.5}
\begin{tabular}{p{0.97\textwidth}}
\toprule
You are a political stance evaluator. Your task is to determine whether the \texttt{Steered Output} successfully transforms the political polarity of the \texttt{Original Input} from left-leaning to right-leaning, while maintaining coherence and relevance. Use the following criteria to guide your evaluation:

1. \textbf{Right-Leaning Position Achieved}: The output expresses a clearly conservative or right-leaning perspective. Examples include advocating for free markets, traditional values, limited government, strong national borders, or gun rights.

2. \textbf{Topical Relevance}: The output remains on-topic, addressing the same subject or theme as the original input, rather than introducing unrelated content.

3. \textbf{Content Richness}: The output should not simply repeat a phrase or contain incoherent text. It must present a clear, meaningful argument or viewpoint.

\textbf{Original Input (Left-Leaning Statement):} \{original\_input\}

\textbf{Steered Output:} \{steered\_output\}

\textbf{Question:} Does the \texttt{Steered Output} meet all three criteria above?

\textbf{Output Format:} Your response MUST start with \texttt{"Yes"} or \texttt{"No"} on the first line. If your answer is \texttt{"No"}, briefly explain why (e.g., ``Still left-leaning'', ``Off-topic'', or ``Repetitive content'').

\textbf{Example for "Yes":} \\
Yes \\
The output opposes government intervention and emphasizes free-market solutions, staying on-topic.

\textbf{Example for "No":} \\
No \\
The output remains left-leaning and calls for more public programs.

\end{tabular}
\end{spacing}
}
\label{tab:polarity_eval_prompt}
\end{table*}

\end{document}